\documentclass[letterpaper, 10 pt, conference]{ieeeconf}  

\IEEEoverridecommandlockouts                              
\usepackage{graphicx}
\usepackage{rotating}
\usepackage{adjustbox}
\usepackage{multirow}
\usepackage{color}
\usepackage[pdftex,dvipsnames]{xcolor}
\usepackage{amsmath}
\usepackage{algorithm}
\usepackage{amssymb}
\usepackage[noend]{algpseudocode}
\usepackage{soul}
\usepackage{bm}
\usepackage{comment}

\DeclareMathOperator*{\argmin}{arg\,min}

\makeatletter
\newcommand*\dotp{\mathpalette\dotp@{.5}}
\newcommand*\dotp@[2]{\mathbin{\vcenter{\hbox{\scalebox{#2}{$\m@th#1\bullet$}}}}}
\makeatother

\makeatletter
\newcommand*{\itemequation}[3][]{%
  \item
  \begingroup
    \refstepcounter{equation}%
    \ifx\\#1\\%
    \else
      \label{#1}%
    \fi
    \sbox0{#2}%
    \sbox2{$\displaystyle#3\m@th$}%
    \sbox4{ \@eqnnum}%
    \dimen@=.5\dimexpr\linewidth-\wd2\relax
    \let\CenterInSpace=N%
    \ifcase
        \ifdim\wd0>\dimen@
          \z@
        \else
          \ifdim\wd4>\dimen@
            \z@
          \else
            \@ne
          \fi
        \fi
      \let\CenterInSpace=Y%
    \fi
    \ifdim\dimexpr\wd0+\wd2+\wd4\relax>\linewidth
      \@latex@warning{Equation is too large}%
    \fi
    \noindent
    \rlap{\copy0}%
    \ifx\CenterInSpace Y%
      \rlap{\hbox to \linewidth{\kern\wd0\hss\copy2\hss\kern\wd4}}%
    \else
      \rlap{\hbox to \linewidth{\hfill\copy2\hfill}}%
    \fi
    \hbox to \linewidth{\hfill\copy4}%
    \hspace{0pt}
  \endgroup
  \ignorespaces
}
\makeatother

\usepackage{svg}
\usepackage{cite}
\usepackage{caption}
\usepackage{subcaption}
\usepackage{dblfloatfix}
\usepackage{siunitx}

\usepackage[pdftex,dvipsnames]{xcolor}
\usepackage[colorinlistoftodos,prependcaption,textwidth=100,textsize=normalsize]{todonotes}

\title{\LARGE \bf
Self-Supervised Traversability Prediction by Learning to Reconstruct Safe Terrain
}

\author{Robin Schmid$^{1, 2}$, Deegan Atha$^{1}$, Frederik Schöller$^{1,4}$, Sharmita Dey$^{1,3}$, Seyed Fakoorian$^{1}$, Kyohei Otsu$^{1}$, \\ Barry Ridge$^{1}$, Marko Bjelonic$^{2}$, Lorenz Wellhausen$^{2}$, Marco Hutter$^{2}$, Ali-akbar Agha-mohammadi$^{1}$%
\thanks{$^{1}$Jet Propulsion Laboratory (JPL), California Institute of Technology (Caltech), Pasadena, CA, United States of America}%
\thanks{$^{2}$Swiss Federal Institute of Technology (ETH Zürich), Robotic Systems Lab, Switzerland
        {\tt\small {schmirob}@ethz.ch}}%
\thanks{$^{3}$University of Goettingen, Department of Computer Science, Germany}%
\thanks{$^{4}$Technical University of Denmark, Department of Electrical Engineering and Photonics, Denmark}%
\thanks{The research was carried out at the Jet Propulsion Laboratory, California Institute of Technology, under a contract with the National Aeronautics and Space Administration (80NM0018D0004), partial funded by the Jet Propulsion Laboratory and partially funded by Swiss Federal Institute of Technology, ETH Zürich.}%
\thanks{\copyright2022. All Rights Reserved}
}
\begin{document}

\maketitle
\thispagestyle{empty}
\pagestyle{empty}


\begin{abstract}
Navigating off-road with a fast autonomous vehicle depends on a robust perception system that differentiates traversable from non-traversable terrain. Typically, this depends on a semantic understanding which is based on supervised learning from images annotated by a human expert. This requires a significant investment in human time, assumes correct expert classification, and small details can lead to misclassification. To address these challenges, we propose a method for predicting high- and low-risk terrains from only past vehicle experience in a self-supervised fashion. First, we develop a tool that projects the vehicle trajectory into the front camera image. Second, occlusions in the 3D representation of the terrain are filtered out. Third, an autoencoder trained on masked vehicle trajectory regions identifies low- and high-risk terrains based on the reconstruction error. We evaluated our approach with two models and different bottleneck sizes with two different training and testing sites with a four-wheeled off-road vehicle. Comparison with two independent test sets of semantic labels from similar terrain as training sites demonstrates the ability to separate the ground as low-risk and the vegetation as high-risk with 81.1\% and 85.1\% accuracy.
\end{abstract}


\section{Introduction}
\begin{figure}[t]
  \centering
  \includegraphics[width=0.48\textwidth]{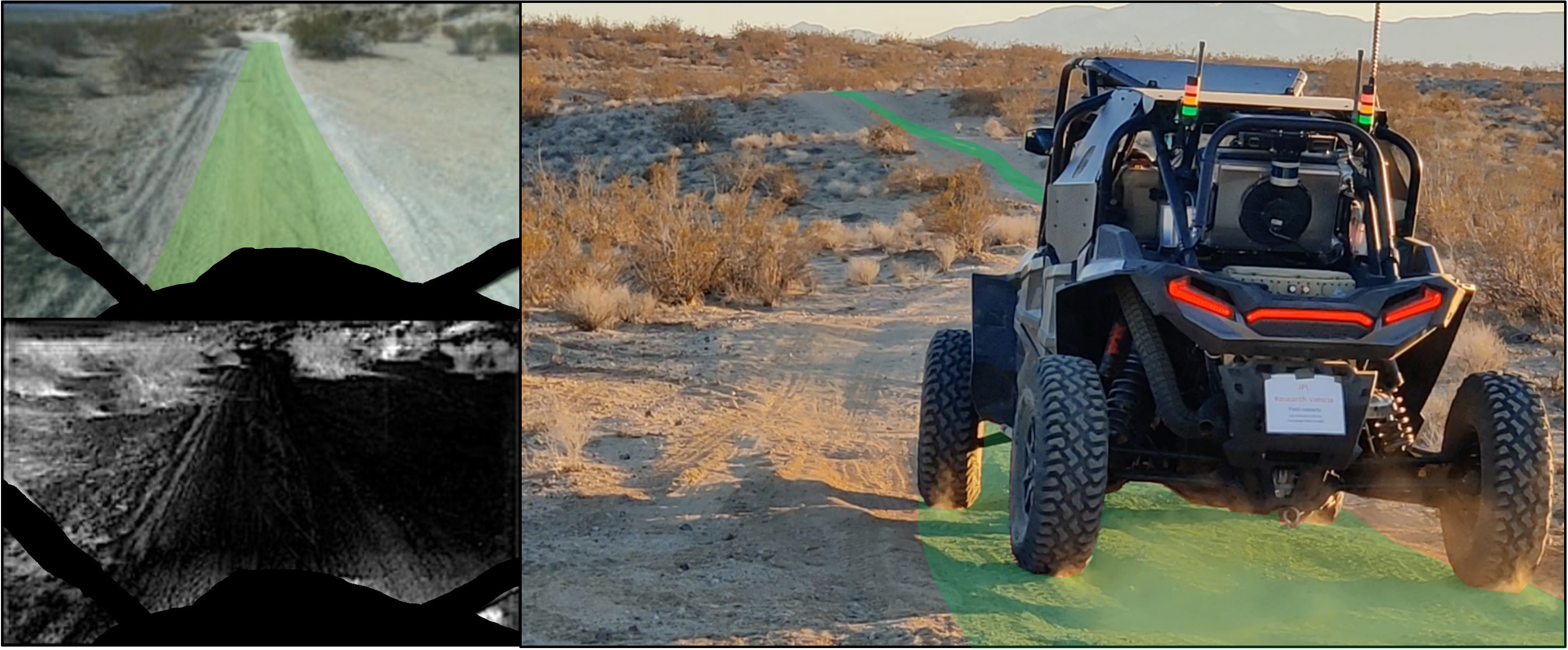}
  \caption{Polaris RZR vehicle autonomy testing in Mojave desert. \textit{Top left image} and \textit{right image}: \textit{Green} indicates the region traversed by the vehicle. \textit{Bottom left image}: Predicted traversable region by the model.}
  \label{fig:front_page}
\end{figure}

Fast autonomous off-road and off-trail driving requires robust and accurate perception and understanding of the unstructured terrain in which the vehicle is navigating. It also often necessitates traversing through different surface types which could include different types of traversable and non-traversable vegetation or surfaces with different properties such as sand or soil. Therefore, it is crucial to understand what surfaces pose a low risk to the vehicle and which areas have a higher risk. However, many geometric-based approaches \cite{Overbye2021GVOMAG, Fankhauser2018ProbabilisticTM} typically require additional terrain classification algorithms based on supervised learning to capture the variety of terrain \cite{Sofman2006ImprovingRN, lalonde, jiang2020rellis3d}. Semantic labeling requires a significant investment of human time to manually annotate the data. Additionally, the boundary between many different classes, especially vegetation types, can be difficult and laborious to determine for human annotators, since some of the largest public data sets contain data from only a single natural environment \cite{yamaha, freiburgforest, jiang2020rellis3d}. Therefore, it would be ideal to learn which terrain in any environment is traversable and which is non-traversable using only previous experiences of the vehicle via a self-supervised approach \cite{Wellhausen2019WhereSI, Zurn2021SelfSupervisedVT, Nava2019LearningLP}. Furthermore, to drive fast and with a highly capable vehicle such as the Polaris RZR (Figure \ref{fig:front_page}) in our case, different semantic classes could pose different risks compared to other vehicles. In order to scale a perception system to handle a wide variety of natural terrains and vehicle risk tolerances, efficient self-supervised learning techniques are needed.

\begin{figure*}[!t]
  \centering
  \includegraphics[width=\linewidth]{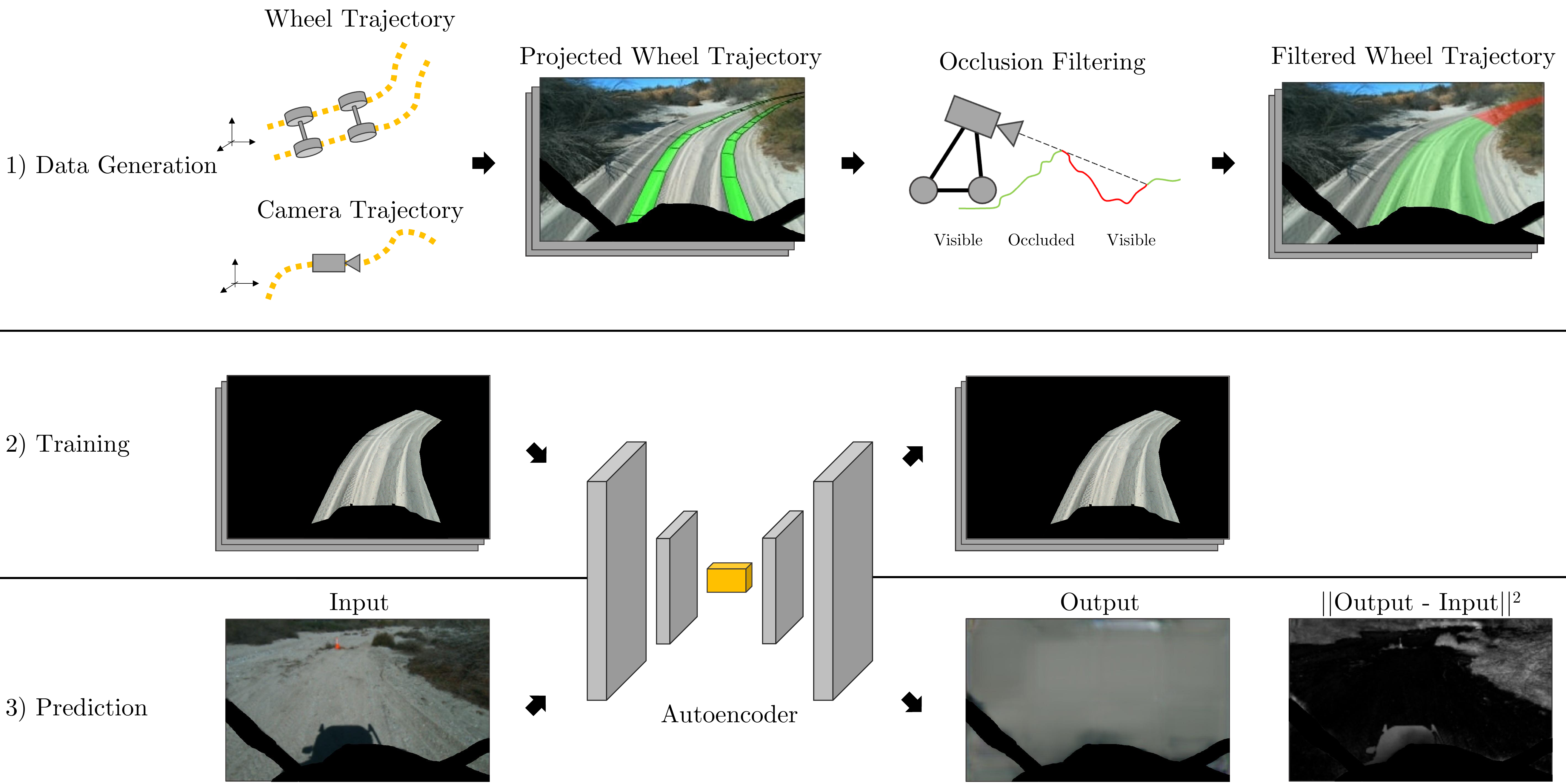}
  \caption{Self-supervised traversability pipeline: 1) In the data generation the wheel trajectory is projected into the front camera image and occluded points are removed from the projected trajectory. 2) A set of training images automatically generated is masked with the projected wheel trajectory and used for the training of an autoencoder. 3) During prediction the front camera image is reconstructed and the reconstruction error is used as a measure of how traversable regions in the image are.}
  \label{fig:overall_architecture}
\end{figure*}
\subsection{Related Work}
Current traversability analysis relies on geometric, semantic, or proprioceptive features
\cite{Papadakis2013TerrainTA, MAARS}. The features depend on the available sensor configuration and vary between robots. Geometric traversability analysis can work for rigid environments and analyzes the terrain based on obstacles, slope, or roughness of the terrain \cite{Overbye2021GVOMAG}. The environment is represented as a 2D, 2.5D \cite{Fankhauser2016AUG} or 3D map \cite{Overbye2021GVOMAG}. Other sensor modalities such as RGB, Near-Infrared (NIR) or RADAR are used to enrich the information used for planning and inferring semantic information.

Learning-based methods for terrain classification have been investigated intensively in recent years. With the success of semantic segmentation models \cite{deeplab}, several data sets and the corresponding supervised terrain segmentation models have been released \cite{yamaha,freiburgforest,jiang2020rellis3d,RUGD2019IROS}. AI4Mars \cite{ai4mars} was able to generate a large labeled data set for the segmentation of Mars terrain, but the level of work required to collect a data set this large is infeasible for many robotic applications. Some such as \cite{freiburgself} utilize both manual labels and self-supervision. However, the corresponding data is focused on one specific environment and still requires manual labels.

The performance of terrain classification significantly depends on the size of the data set. Self-supervised methods deal with this by leveraging data from past experience of the robot. Recent work deployed self-supervised methods for predicting terrain properties at distance from data close to the robot. Proprioceptive data at future time instances were associated with visual images in the data set to predict proprioceptive information from underneath the robot at a farther distance using images. This approach was studied using IMU \cite{Otsu2016AutonomousTC, Bai2019ThreeDimensionalVT}, force-torque sensors \cite{Wellhausen2019WhereSI}, or acoustic signals \cite{Zurn2021SelfSupervisedVT} for the generation of proprioceptive data. In addition, \cite{Nava2019LearningLP} took advantage of this by using a proximity sensor to learn the traversability at distance. Whereas these approaches do not handle occlusions during the labeling process, \cite{Barnes2017FindYO} dealt with trajectories that were partially occluded.

Traversability classification algorithms are prone to overconfident predictions on out-of-distribution samples.
Additionally, negative samples are difficult to collect as they might lead to catastrophic damage to the system.
Autoencoders do not have these issues because they can learn the appearance of previously traversed terrain, only from positive samples.
Therefore, image regions with high reconstruction error are likely to be novelties.
Autoencoders are used for the detection of non-traversable regions for a quadrupedal robot \cite{Wellhausen2020SafeRN}, planetary exploration \cite{Kerner2020ComparisonON} or autonomous driving \cite{Stocco2020MisbehaviourPF, Gu2020ANL}.


\subsection{Contribution}
In this paper, we work towards the goal of self-supervised perception of traversability by using the paths a robot previously successfully traversed in order to learn traversable regions for future navigation. This is accomplished via a developed projection tool that projects the wheel positions of the vehicle's future path into 2D images at previous timestamps. This pipeline follows a similar approach \cite{Wellhausen2019WhereSI} and extends it in several key areas. First, the tool utilizes multiple LIDAR scans at a single instance for 3D representation of the terrain and second, it has the ability to filter occluded regions that are prevalent within off-trail environments. This tool generates 2D trajectory labels which we use to train a model to predict which regions have low traversability risk and which have high traversability risk. We demonstrate that an autoencoder trained on the masked trajectory region can identify low and high-risk terrains via differences in predicted reconstruction error.

The key contributions can be summarized as follows: 
\begin{itemize}
\item Creation of a wheel projection tool utilizing a 3D world model from multiple LIDAR scans
\item Occlusion filtering for trajectory masks
\item An autoencoder model that can predict high- and low-risk terrains based on the wheel projection labels
\end{itemize}


\section{Method}
\begin{figure*}[!t]
\centering
\begin{subfigure}{0.24\textwidth}
  \centering
  \includegraphics[width=\linewidth]{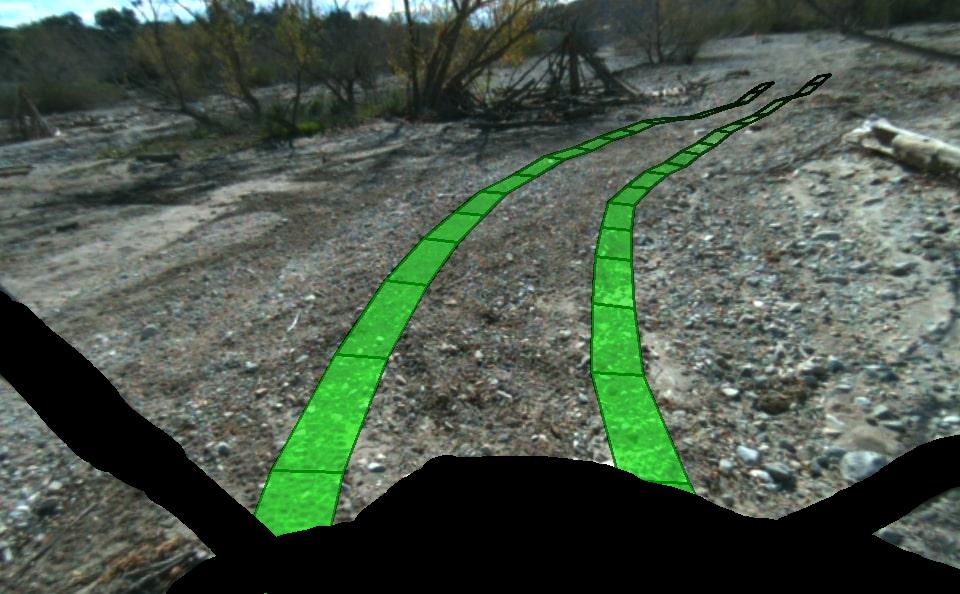}
  \caption{}
\end{subfigure}
\begin{subfigure}{0.24\textwidth}
  \centering
  \includegraphics[width=\linewidth]{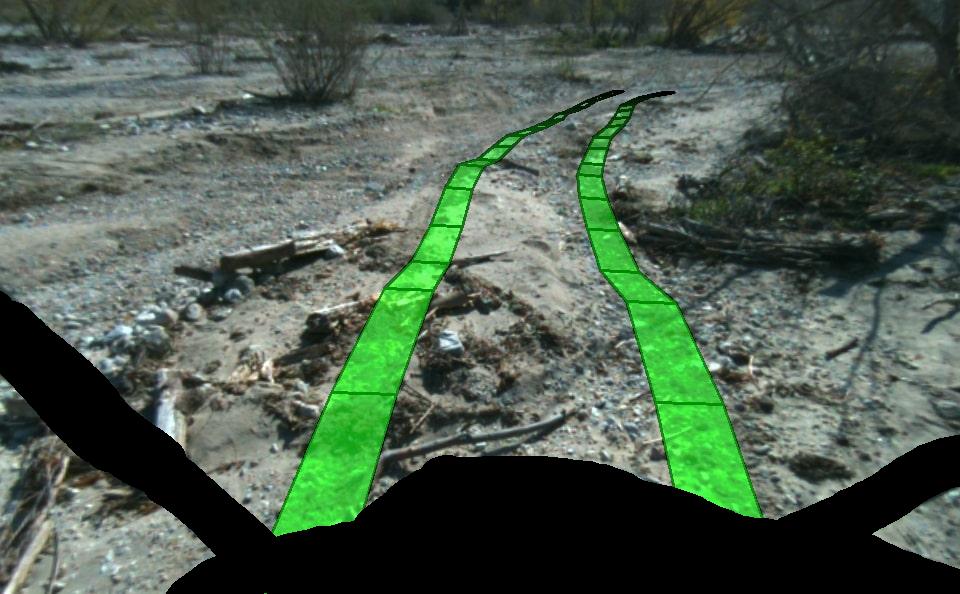}
  \caption{}
\end{subfigure}
\begin{subfigure}{0.24\textwidth}
  \centering
  \includegraphics[width=\linewidth]{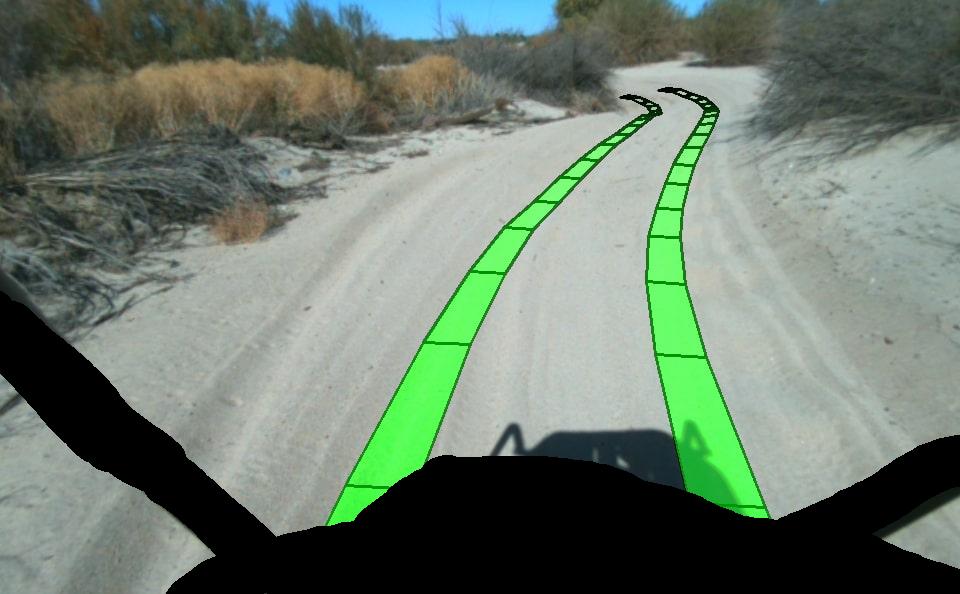}
  \caption{}
\end{subfigure}
\begin{subfigure}{0.24\textwidth}
  \centering
  \includegraphics[width=\linewidth]{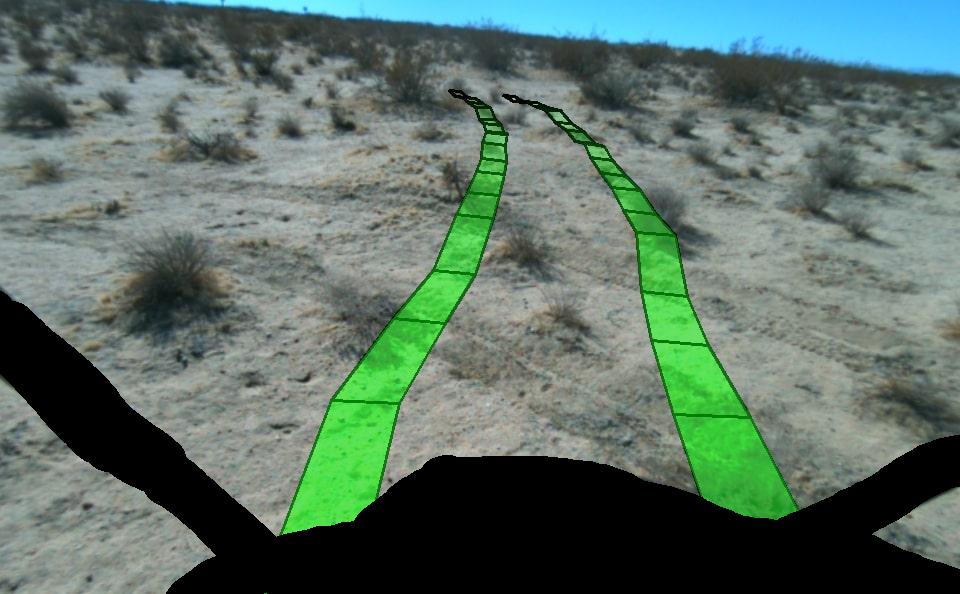}
  \caption{}
\end{subfigure}
\caption{Data collection on different terrain: The vehicle trajectory is projected onto the front camera image. a) Arroyo with gravel terrain, b) Arroyo with wood logs and rocks, c) Mojave with a sand trail, d) Mojave with hardpack and vegetation.}
\label{fig:test_sites}
\end{figure*}

Our approach for self-supervised traversability detection is shown in Figure \ref{fig:overall_architecture}. To generate training labels the wheel positions are projected into the camera images and filtered for occlusions. The masks generated are used to train an autoencoder that learns to represent parts of the image within this mask. During inference, we use the reconstruction error of the autoencoder to determine traversable regions.

\subsection{Projecting Wheel Tracks to Camera Image}
In order to project the wheel positions into an image of the camera, their positions with respect to the camera need to be known. First, we need to find the pose of the vehicle in the global frame. In order to do so, the open-source state estimation framework, LIO-SAM \cite{Shan2020LIOSAMTL} is used, which provides a low-drift pose estimate at every timestamp with respect to the body-frame (base link). We can then obtain the wheel contacts from the base link coordinate through a static transform. Therefore, for each of the two front wheels, we compute the position of two contact points on either side of the wheel with the ground and assume that the wheel contact region is a line of length of the wheel width.

At each time instance $t$ at which an image from the front camera is captured, the contact points of the wheel in the global frame $p^t \in \mathbb{R}^{3}$ and the time-dependent transformation from the wheel to the camera $T_{cw}^t$ are stored. Denoting the camera intrinsic calibration matrix $K$ and the extrinsic calibration matrix $P$ the wheel image points in homogeneous coordinates $i^t =$ $\begin{bmatrix} u & v & 1 \end{bmatrix}^T$ are computed with
\begin{align}
    i^t = P K T_{cw}^t p^t .
\end{align}
This projection is computed from $t$ to $t+\tau$ with $\tau$ as the projection horizon, which projects the wheel trajectory $p^{t:t+\tau}$ to the wheel image points $i^{t:t+\tau}$. These points are then connected to a quadrilateral resulting in the connected trajectory in Figure \ref{fig:test_sites}.

\subsection{Occlusion Filtering}
The terrain on which we operate can be unstructured and contain obstacles such as boulders, vegetation, or ditches. Since the goal is to learn from pairs of visual and proprioceptive data, only parts in the image which represent the corresponding ground patch of the collected data should be visible. Parts of the trajectory behind an occlusion may yield misleading information and need to be filtered. This occlusion filtering takes into account the geometric information in form of a point cloud.

At time instance $t$ at which the image is taken the point cloud $c^t$ is projected into image coordinates. The wheel points trajectory and point cloud are then converted into spherical coordinates. A potential occlusion point $o$ that occludes a point on the wheel is found by a nearest neighbor search in the azimuthal and radial dimension for each wheel point over all points of the point cloud. This finds the point in the point cloud that is closest to the ray from the camera to the wheel point.

A wheel point is treated as occluded if the relative radial distance of this potential occluded wheel point is less than a radial distance threshold $\rho$. This threshold allows to adjust the size of the obstacles filtered out. The occluded points are then removed from the wheel trajectory.

\begin{algorithm}[H]\label{alg:occlusion}
\caption{Occlusion Filtering}
\begin{algorithmic}
\Statex $\textbf{Initialization: } p^{t:t+\tau}, c^{t}$
\For{$p$ in $p^{t:t+\tau}$}
\State $o \gets \text{nearestNeighbor}([p_{\theta}, p_{\phi}], [c_{\theta}^{t}, c_{\phi}^{t}])$
\If{$\frac{o_r - p_r}{p_r} < \rho$}
\State $p \gets \text{is occluded}$
\Else{}
\State $p \gets \text{is not occluded}$
\EndIf
\EndFor
\end{algorithmic}
\end{algorithm}

\subsection{Traversability Learning} \label{meth:learning}
\begin{figure*}
\centering
\begin{subfigure}{0.24\textwidth}
  \centering
  \includegraphics[width=\linewidth]{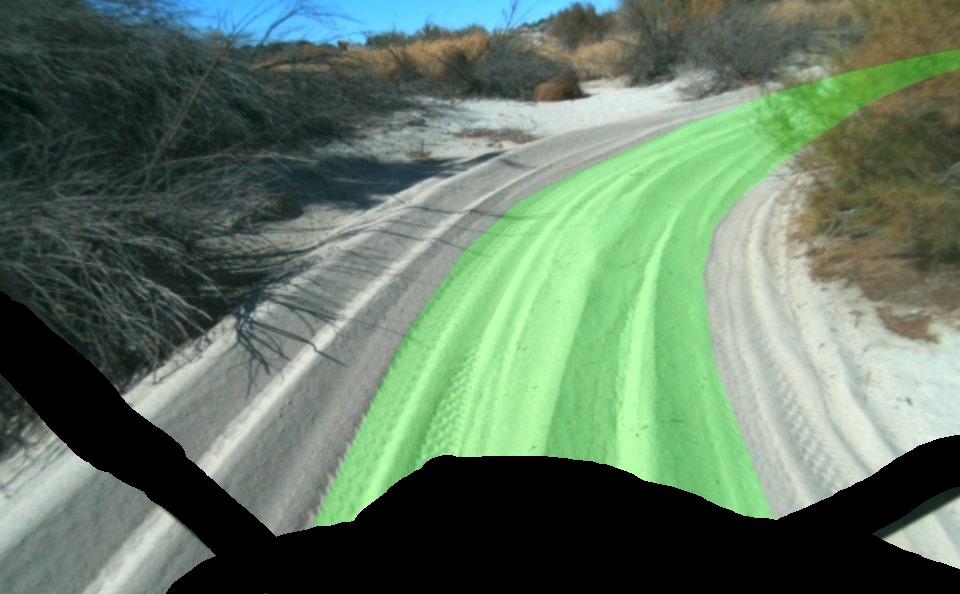}
\end{subfigure}
\begin{subfigure}{0.24\textwidth}
  \centering
  \includegraphics[width=\linewidth]{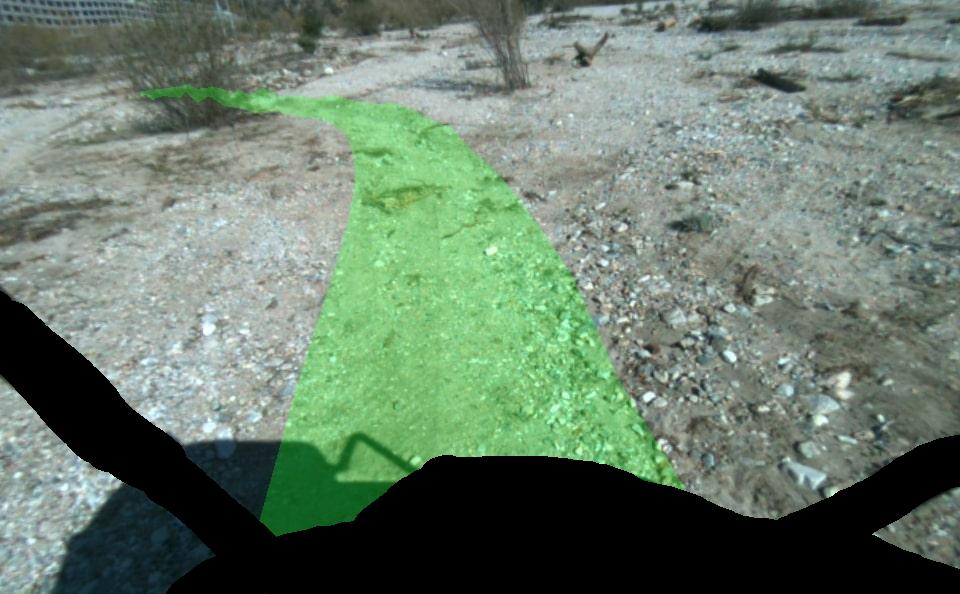}
\end{subfigure}
\begin{subfigure}{0.24\textwidth}
  \centering
  \includegraphics[width=\linewidth]{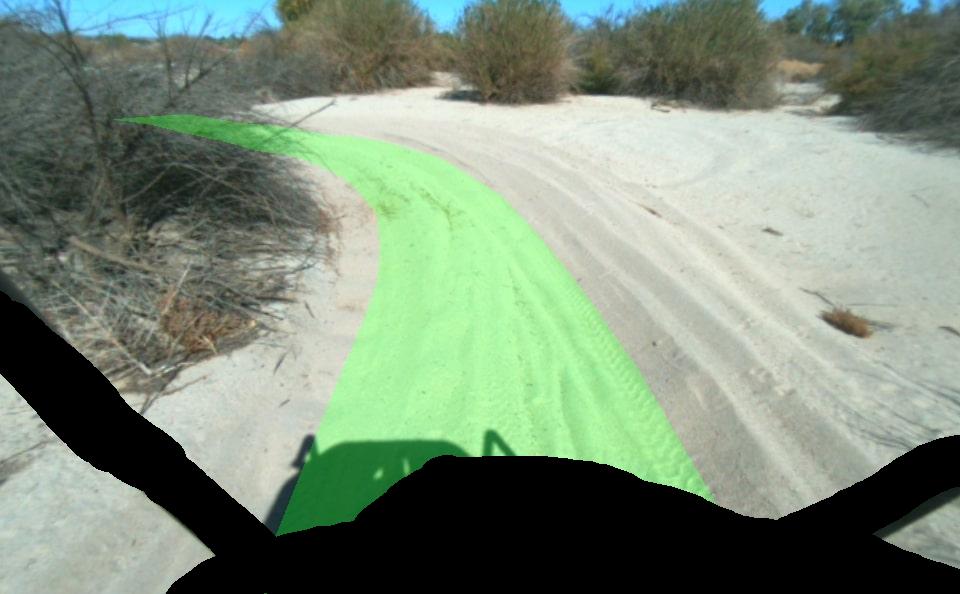}
\end{subfigure}
\begin{subfigure}{0.24\textwidth}
  \centering
  \includegraphics[width=\linewidth]{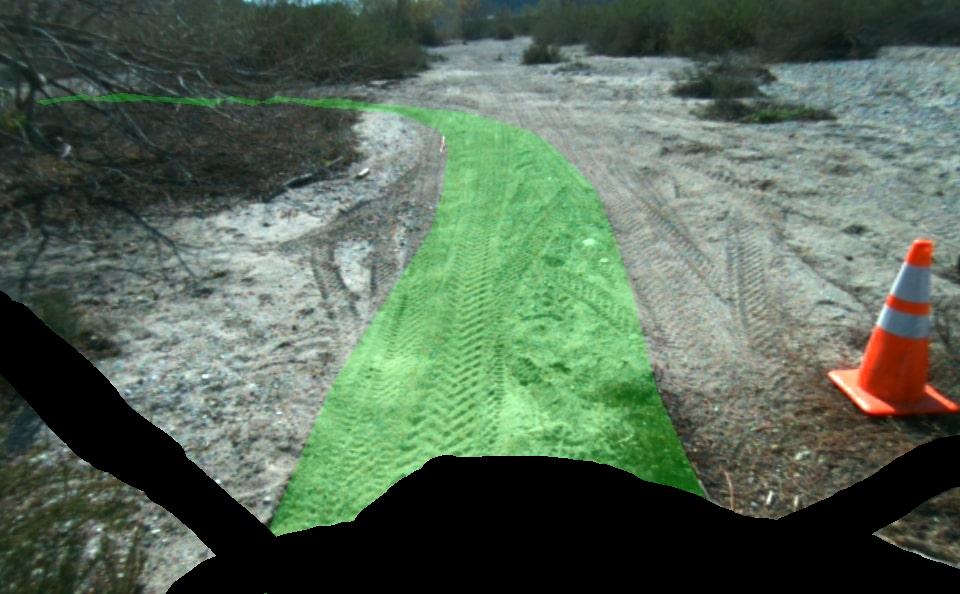}
\end{subfigure}
\bigskip
\begin{subfigure}{0.24\textwidth}
  \centering
  \includegraphics[width=\linewidth]{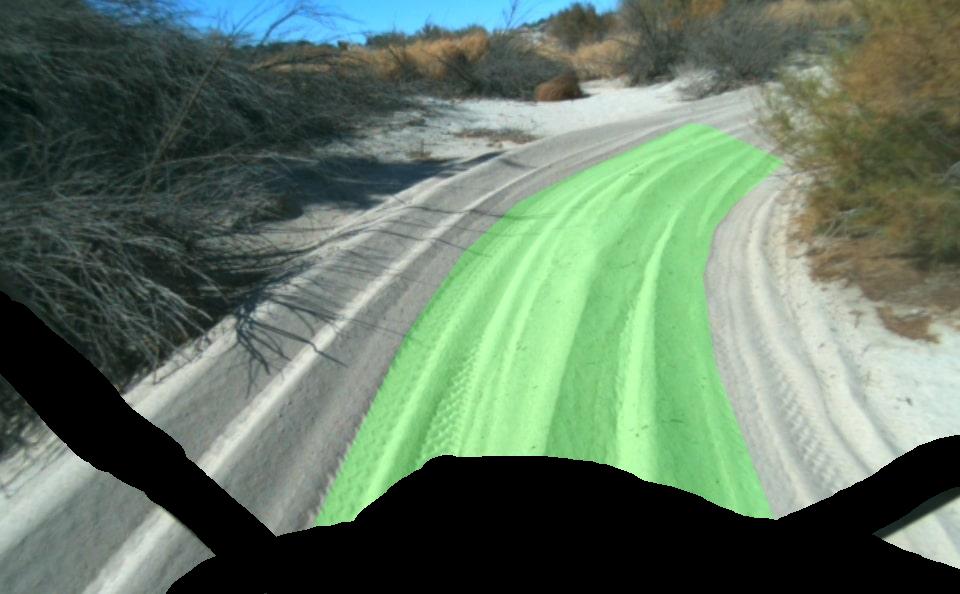}
\end{subfigure}
\begin{subfigure}{0.24\textwidth}
  \centering
  \includegraphics[width=\linewidth]{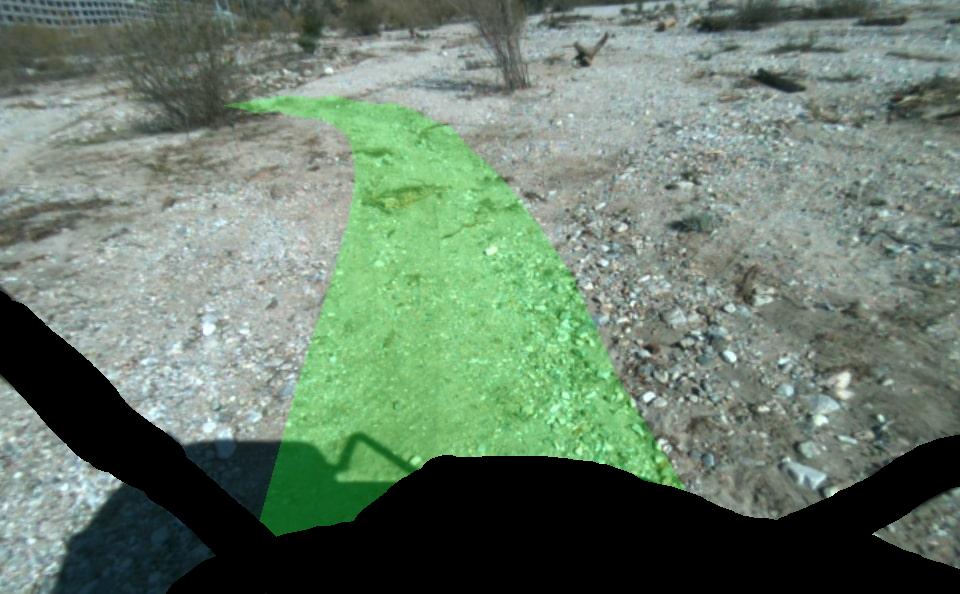}
\end{subfigure}
\begin{subfigure}{0.24\textwidth}
  \centering
  \includegraphics[width=\linewidth]{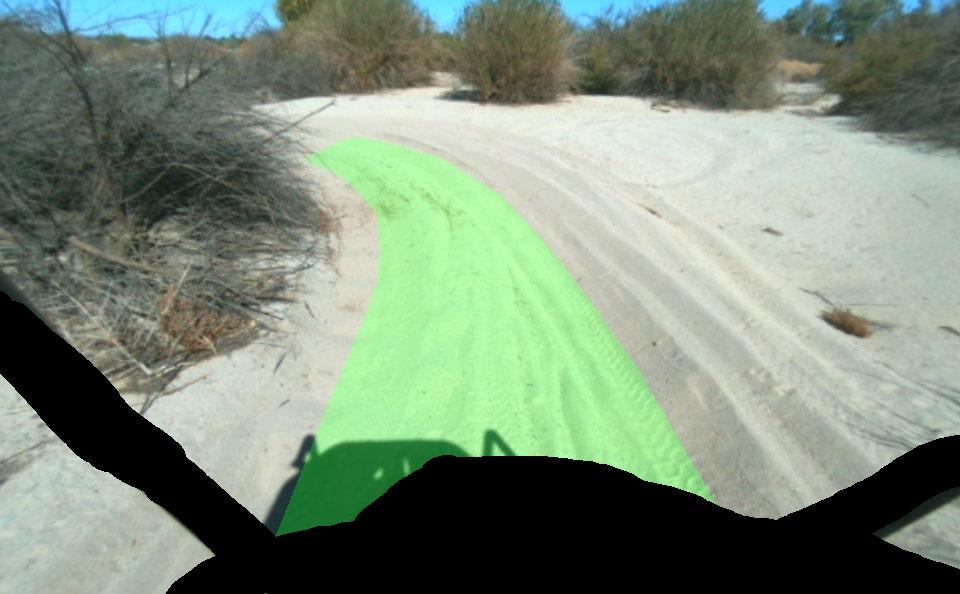}
\end{subfigure}
\begin{subfigure}{0.24\textwidth}
  \centering
  \includegraphics[width=\linewidth]{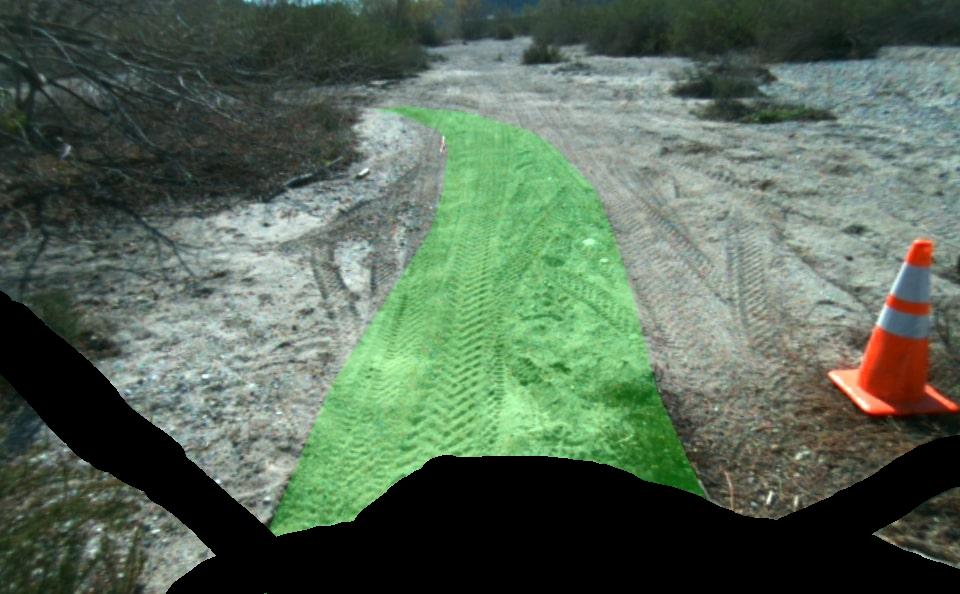}
\end{subfigure}
\caption{Underfoot projections without and with occlusion filtering. The parts of the trajectory which are occluded by vegetation are removed using a 3D representation of the scene.}
\label{fig:occlusion_filtering}
\end{figure*}

In order to predict whether a region is traversable, we use an autoencoder model. The model is optimized against the mean squared error (MSE) loss between an input image, $x$ and its reconstruction, $\hat{x}$. The loss is multiplied element-wise with a binary mask, $m$, that is, 1 within the trajectory region and 0 outside the region and on vehicle parts
\begin{align}
        \mathcal{L}(\hat{x}) = \frac{1}{wh}\sum_{i=0}^{w}\sum_{j=0}^{h} m_{i,j}(\hat{x}_{i,j}-x_{i,j})^2 \quad
\end{align}
where $w$, $h$ is the width and height of the input image $x$. Therefore, the region outside of the masked region is ignored during loss calculation. Using this approach, the reconstruction error will be minimized for regions that have been successfully traversed only. During inference, areas within the model output that have a large reconstruction error are unlikely to have been seen within the training set of traversable regions. These high reconstruction error regions are considered high-risk terrain and low reconstruction error regions correspond to low-risk terrain.  

We use a standard variational autoencoder with Resnet \cite{resnet} backbone to evaluate this approach. The model contains an encoder that takes an image as input and compresses it into a latent space consisting of a $n$-dimensional mean and variance vector, $n$ being the size of the bottleneck layer. The decoder then attempts to reconstruct the image from the latent vector. 



\section{Experiments and Results}

\subsection{Label Generation Details}
\subsubsection{Data Generation}
The data was collected from the four-wheeled Polaris S4 1000 RZR platform. This rugged, autonomous-ready off-road vehicle is equipped with various sensors including RGB stereo cameras and 3x LIDARs (Velodyne VLP-32C). We collected data from different test sites in the Arroyo Seco near the Jet Propulsion Lab in Pasadena, California, and the Mojave Desert. The data contains different terrain types such as gravel, small bushes, sand, waterbed and logs as shown in Figure \ref{fig:test_sites}. Vegetation, slopes, and boulders generate positive obstacles and need to be filtered out. In total, 4000 images of size 960x594 were collected with 2000 in the Arroyo Seco and 2000 in the Mojave Desert, which corresponds to a driving distance per site of around 20 km.

\subsubsection{Point Cloud Filters}\label{exp:point_cloud}
To generate a geometric 3D representation of the terrain, three LIDARs are used. The processed point clouds run through the filtering and merging pipeline developed in previous work \cite{Fan2021STEPST}. The point clouds are spatially merged and dust particles are filtered. For consistency, spacial merging is performed. The points are then segmented into surface, obstacle and ground class, from which outlier points are removed to build a smooth ground surface.

\subsubsection{Wheel Trajectory Generation}
The wheel points are projected at a rate of \SI{10}{\hertz}, \SI{4}{\second} ahead of the robot. Maintaining an average constant driving speed of \SI{30}{\kilo\meter/\hour} results in trajectories of around \SI{35}{\meter} without removing occluded parts. This covers the front camera image to a large extent.

The occlusion filtering was tested on unstructured terrain with positive obstacles such as the vegetation shown in Figure \ref{fig:occlusion_filtering}. The radial distance threshold $\rho$ is determined empirically to 0.35 based on the size of the obstacles present. The number of occlusions on the wheel trajectory increases significantly with a longer wheel projection horizon.

\subsection{Training}
To evaluate the performance of the autoencoder, the model is trained on both the Arroyo and Mojave data sets separately. These data sets were sampled randomly with an 80\% training and 20\% validation split. The models are trained for 100 epochs and are saved based on validation error. All models are trained with a learning rate of $10^{-4}$, a batch size of 4 and an image size of 224x224.

\begin{table*}[t!]
    \centering
    \caption{Percent intersection between high- and low-risk predictions and ground truth semantic labels of ground and vegetation.}
    \begin{tabular}{|c|c|c|c|c|c|c|c|c|}
    \hline
    \textbf{Model} & \textbf{Bottleneck} & \multicolumn{1}{c|}{\textbf{Train}} & \multicolumn{3}{c|}{\textbf{Arroyo}} & \multicolumn{3}{c|}{\textbf{Mojave}} \\\cline{4-9}
    & & \textbf{Dataset} & \textbf{Ground \%} & \textbf{Vegetation \%} & \textbf{AUROC} & \textbf{Ground \%} & \textbf{Vegetation \%} & \textbf{AUROC} \\
    \hline
    Resnet18 & 256 & Arroyo & 78.6 & 74.7 & 0.825 & 58.2 & 48.7 & 0.541\\
    Resnet18 & 512 & Arroyo & 78.7 & 75.4 & 0.826 & 59.9 & 56.6 & 0.596\\
    Resnet50 & 256 & Arroyo & 78.7 & 74.6 & 0.821 & 58.2 & 48.4 & 0.541\\
    Resnet50 & 512 & Arroyo & 76.6 & 74.8 & 0.816 & 57.6 & 49.4 & 0.541\\
    Resnet18 & 256 & Mojave & 82.6 & 79.9 & 0.860 & \textbf{67.8} & 74.6 & 0.730\\
    Resnet18 & 512 & Mojave & \textbf{85.1} & \textbf{81.1} & \textbf{0.888} & 67.1 & \textbf{77.1} & 0.737\\
    Resnet50 & 256 & Mojave & 78.2 & 79.9 & 0.730 & 67.2 & 76.4 & \textbf{0.760}\\
    Resnet50 & 512 & Mojave & 84.5 & 79.2 & 0.840 & 63.1 & 76.0 & 0.712\\
    \hline
    \end{tabular}
    \label{tab:semantic-results}
\end{table*}

\subsection{Evaluation Against Semantic Labels}
\begin{figure}[t]
\centering
\begin{subfigure}{0.23\textwidth}
  \centering
  \includegraphics[width=\linewidth]{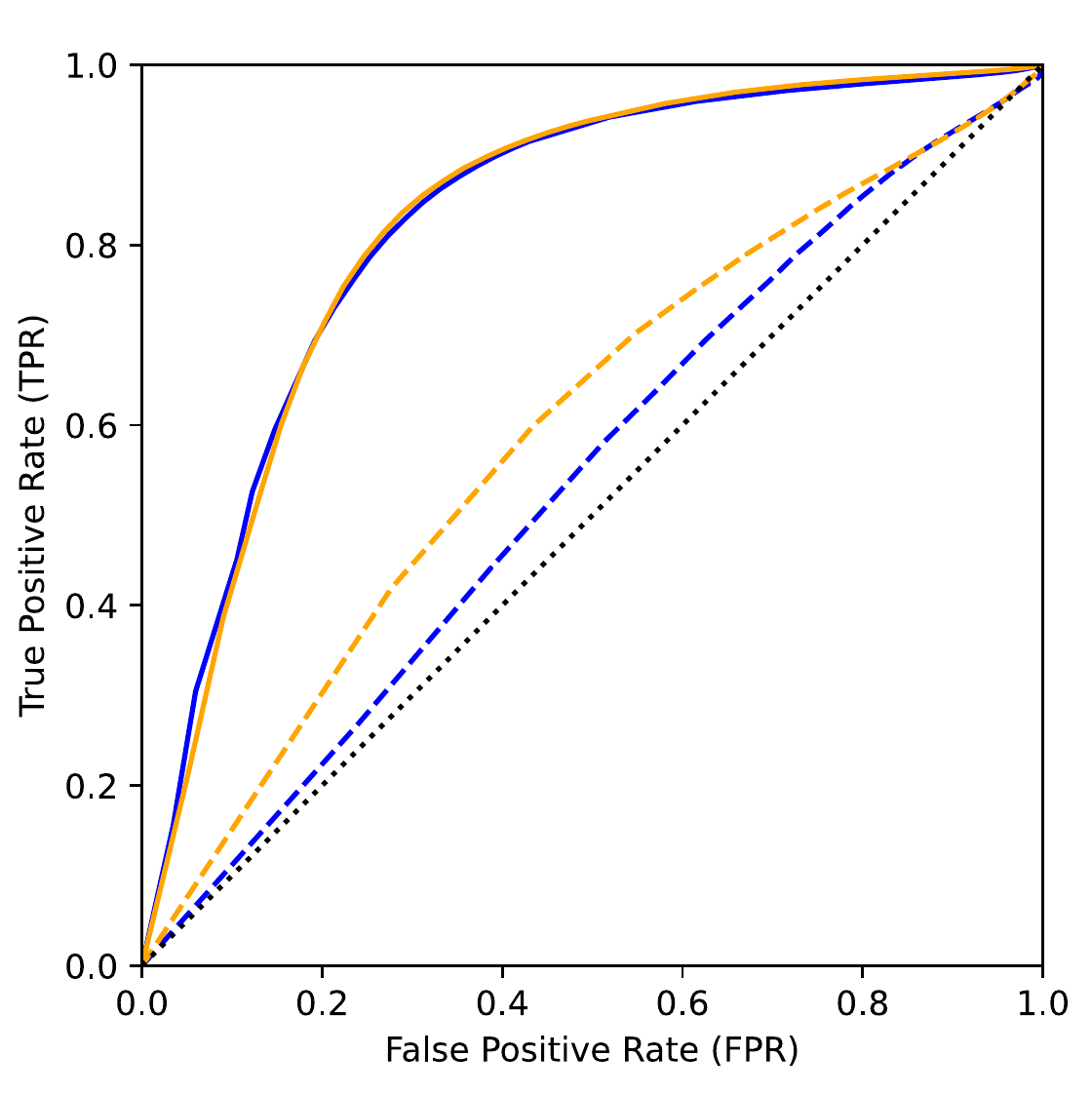}
  \caption{Arroyo trained, Resnet18}
  \label{fig:plot-1}
\end{subfigure}
\begin{subfigure}{0.23\textwidth}
  \centering
  \includegraphics[width=\linewidth]{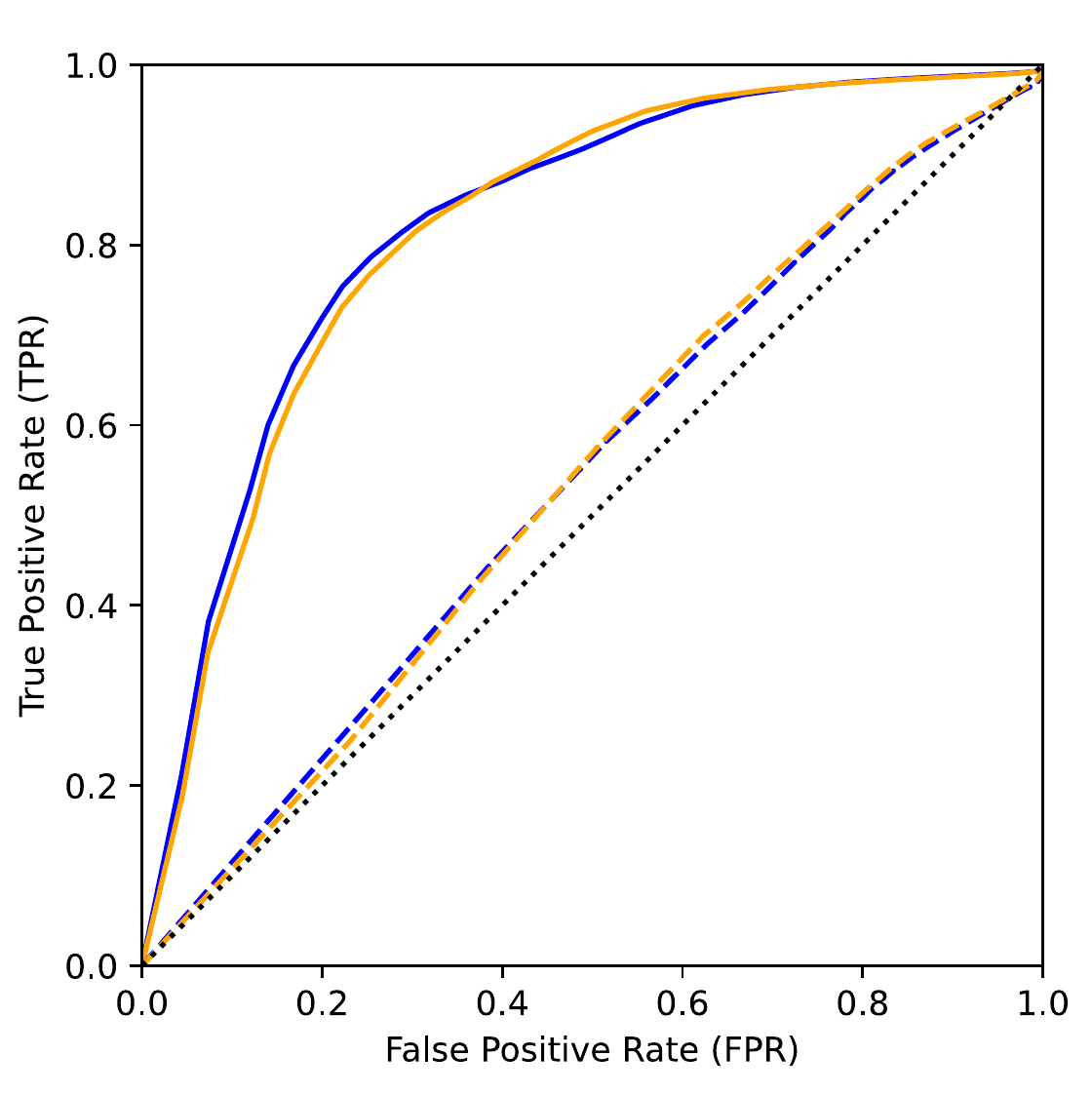}
  \caption{Arroyo trained, Resnet50}
  \label{fig:plot-2}
\end{subfigure}
\begin{subfigure}{0.23\textwidth}
  \centering
  \includegraphics[width=\linewidth]{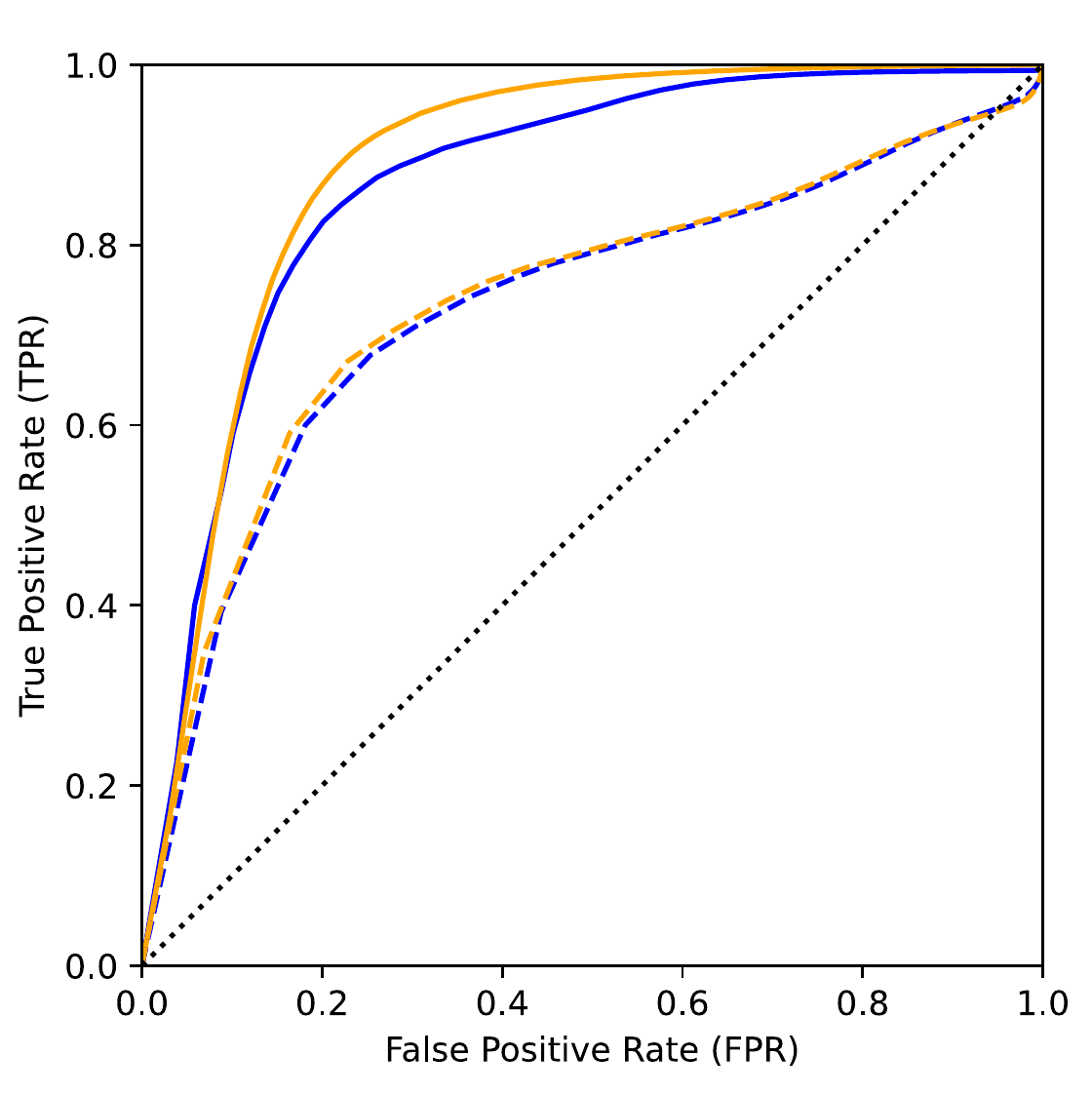}
  \caption{Mojave trained, Resnet18}
  \label{fig:plot-3}
\end{subfigure}
\begin{subfigure}{0.23\textwidth}
  \centering
  \includegraphics[width=\linewidth]{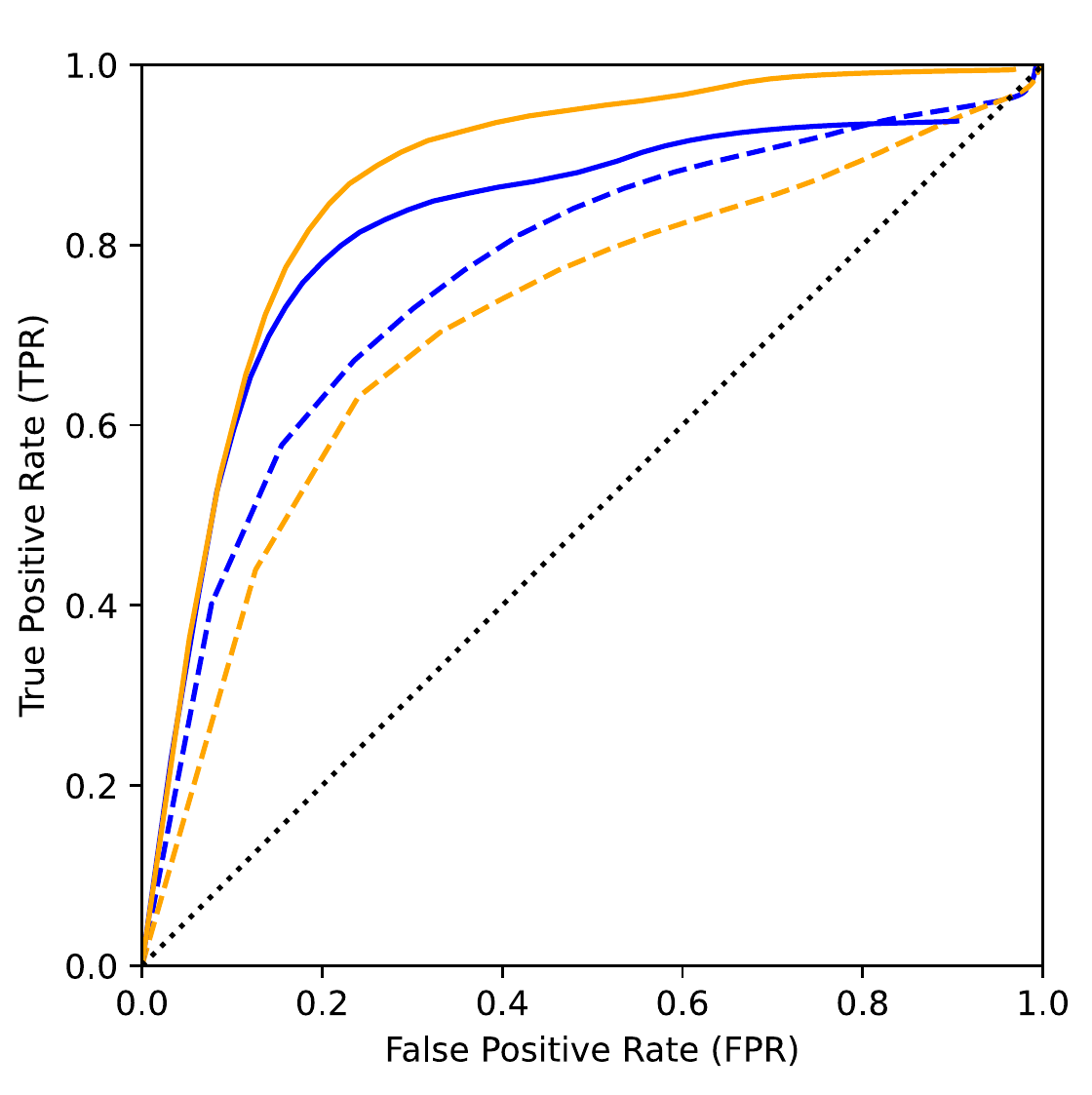}
  \caption{Mojave trained, Resnet50}
  \label{fig:plot-4}
\end{subfigure}
\caption{ROC plots for the Resnet18 and Resnet50 model trained on Arroyo and Mojave data with 256 (\textit{blue}) and 512 (\textit{orange}) bottleneck sizes evaluated on Arroyo (\textit{solid line}) and Mojave (\textit{dashed line}).}
\label{fig:roc-plots}
\end{figure}

In order to further test the performance of the models, each model is compared to independent test sets from Mojave and Arroyo. The Mojave data set contains 955 labeled images of size 960x594 with ground and vegetation segmented. The ground class is a collection of different flat surface terrains, mainly soil and gravel. This data set was collected with our Polaris RZR in separate GPS locations to maintain the independence of the test set. The Arroyo test data contains 1816 labeled images of size 640x480 with ground and vegetation segmentations and is from the MAARS project \cite{MAARS}. This data set was collected at a location similar to our Arroyo training data; however, the images are from an Intel RealSense and from a different season. The autoencoder is evaluated with two different backbone sizes and bottleneck dimensions.
The metric for comparison is the MSE between risk class prediction and semantic class normalized between 0 and 1. This gives a measure where 0 is low risk and 1 is high risk. The receiver operating characteristic curves (ROC) for this experiment are shown in Figure~\ref{fig:roc-plots}. For the Arroyo data set, 0.106 and for Mojave, 0.436 are chosen as a threshold $\theta^{*}$ for low- and high-risk regions for comparison to semantic labels. This threshold is found by optimizing the true positive rate (TPR) and false positive rate (FPR):
\begin{align}
    \theta^{*} = \argmin(\sqrt{(1 - TPR)^{2} + FPR^{2}})
\end{align}

\pagebreak
The overall test results are presented in Table \ref{tab:semantic-results}. Each model, regardless of the training set, is tested on both test sets to assess for overfitting. The Mojave trained model generalizes well, demonstrated by its similar performance on the two test sets. However, the Arroyo trained model shows a lower performance on the Mojave test set. While these semantic labels are not a direct match to traversability risk since they do not capture the vehicle's capabilities, they are a good approximation, as vegetation ideally would have higher risk regardless, due to potential unknown hazards and tire puncture risk. Interestingly, the Arroyo has more non-traversable vegetation and the models predict a higher percentage of this vegetation as high-risk compared to the Mojave data set which contains more traversable vegetation. One of the primary sources of error within the predictions are the shadowed regions on the ground that are predicted with around 60\% as high-risk.\\
Additionally, we observe that in this case the use of a bigger bottleneck size had slightly better results. Due to the size of our training data set, there is likely some overfitting to the specific training terrain using a larger model. Furthermore, using a deeper model, such as the Resnet50, does not appear to have a strong impact on the results.

\begin{figure}[t]
\centering
\begin{subfigure}{0.23\textwidth}
  \centering
  \includegraphics[width=\linewidth]{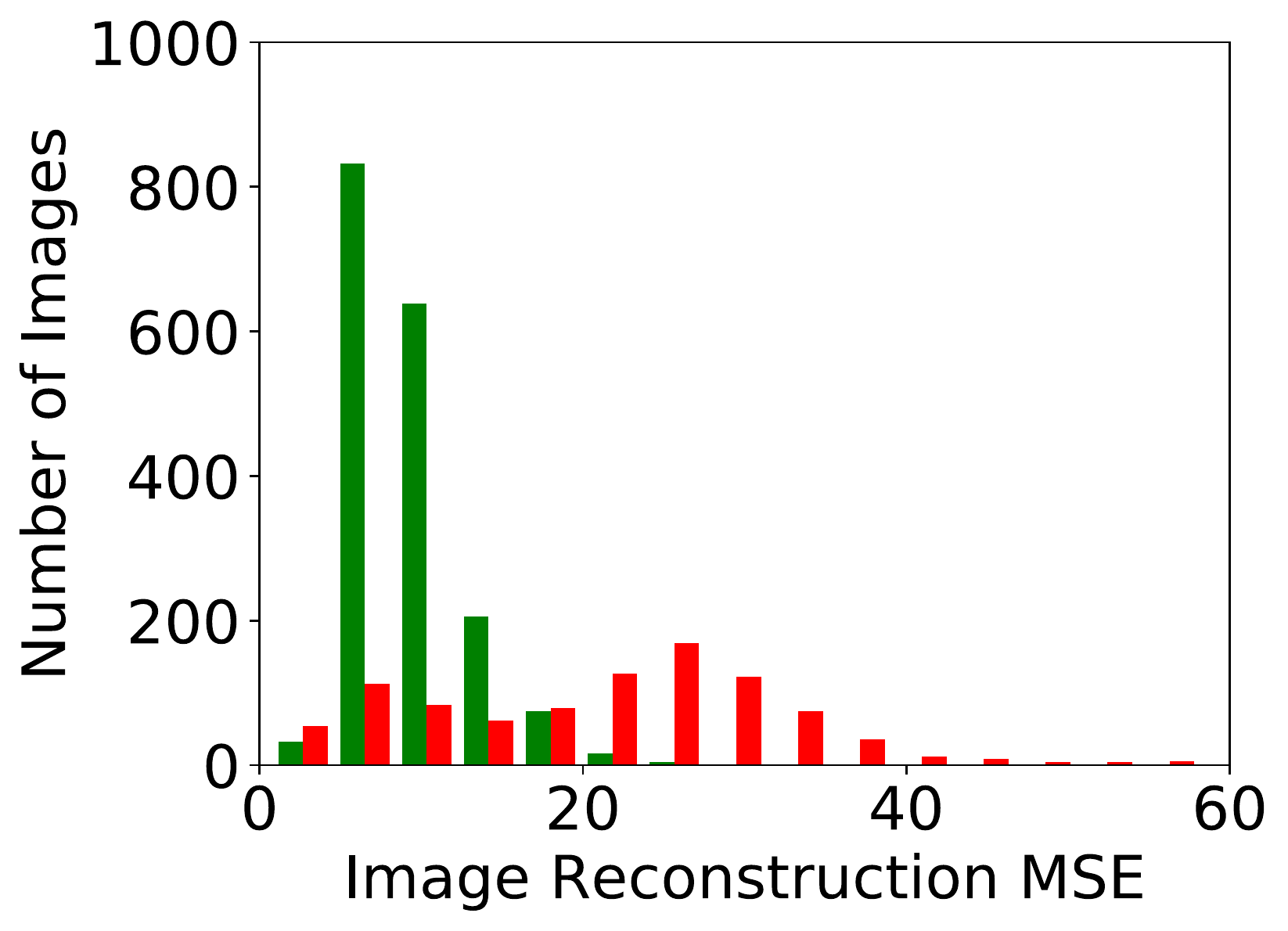}
  \caption{Arroyo model, Arroyo test}
  \label{fig:hist-sem1}
\end{subfigure}
\begin{subfigure}{0.23\textwidth}
  \centering
  \includegraphics[width=\linewidth]{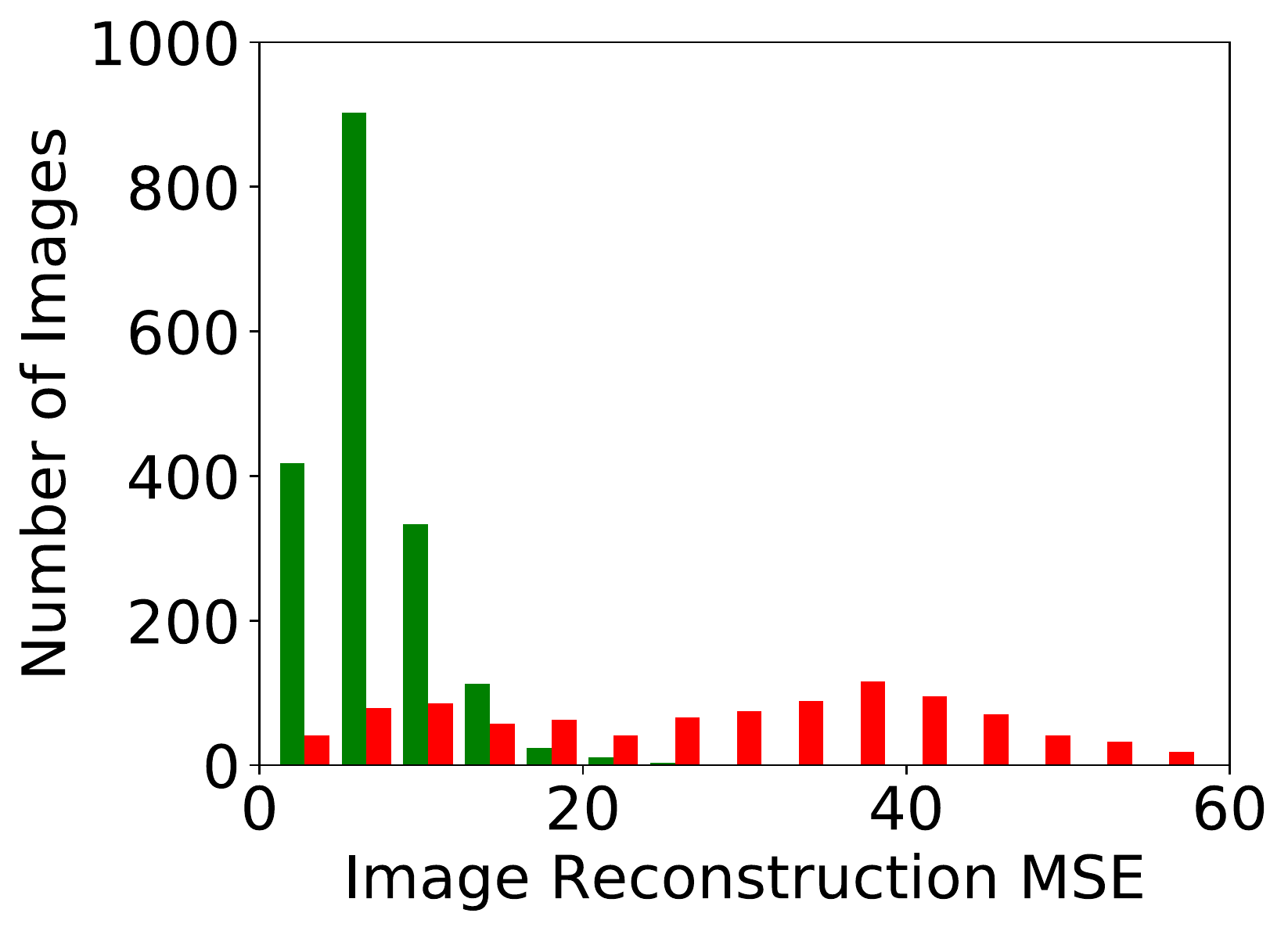}
  \caption{Mojave model, Arroyo test}
  \label{fig:hist-sem2}
\end{subfigure}
\begin{subfigure}{0.23\textwidth}
  \centering
  \includegraphics[width=\linewidth]{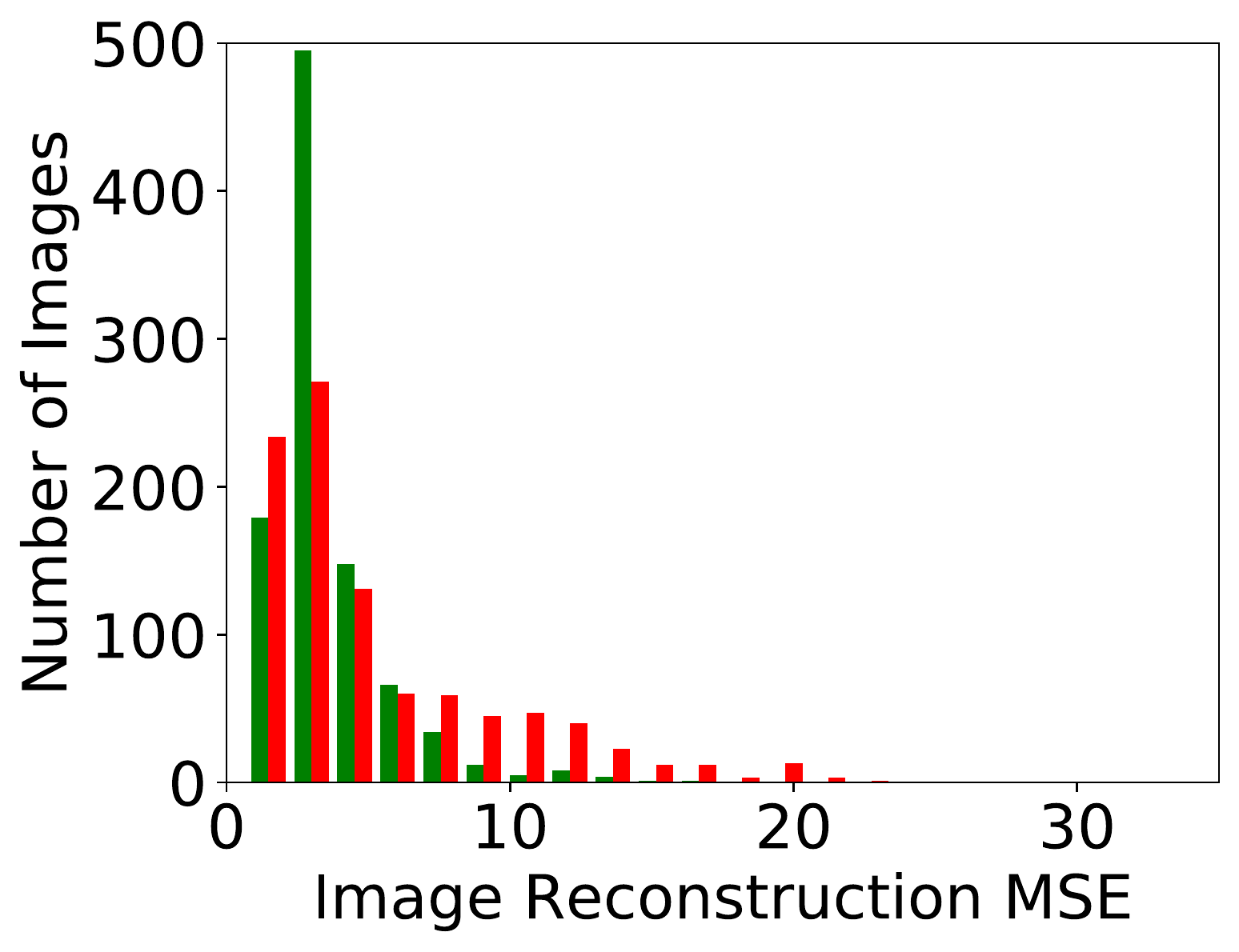}
  \caption{Arroyo model, Mojave test}
  \label{fig:hist-sem3}
\end{subfigure}
\begin{subfigure}{0.23\textwidth}
  \centering
  \includegraphics[width=\linewidth]{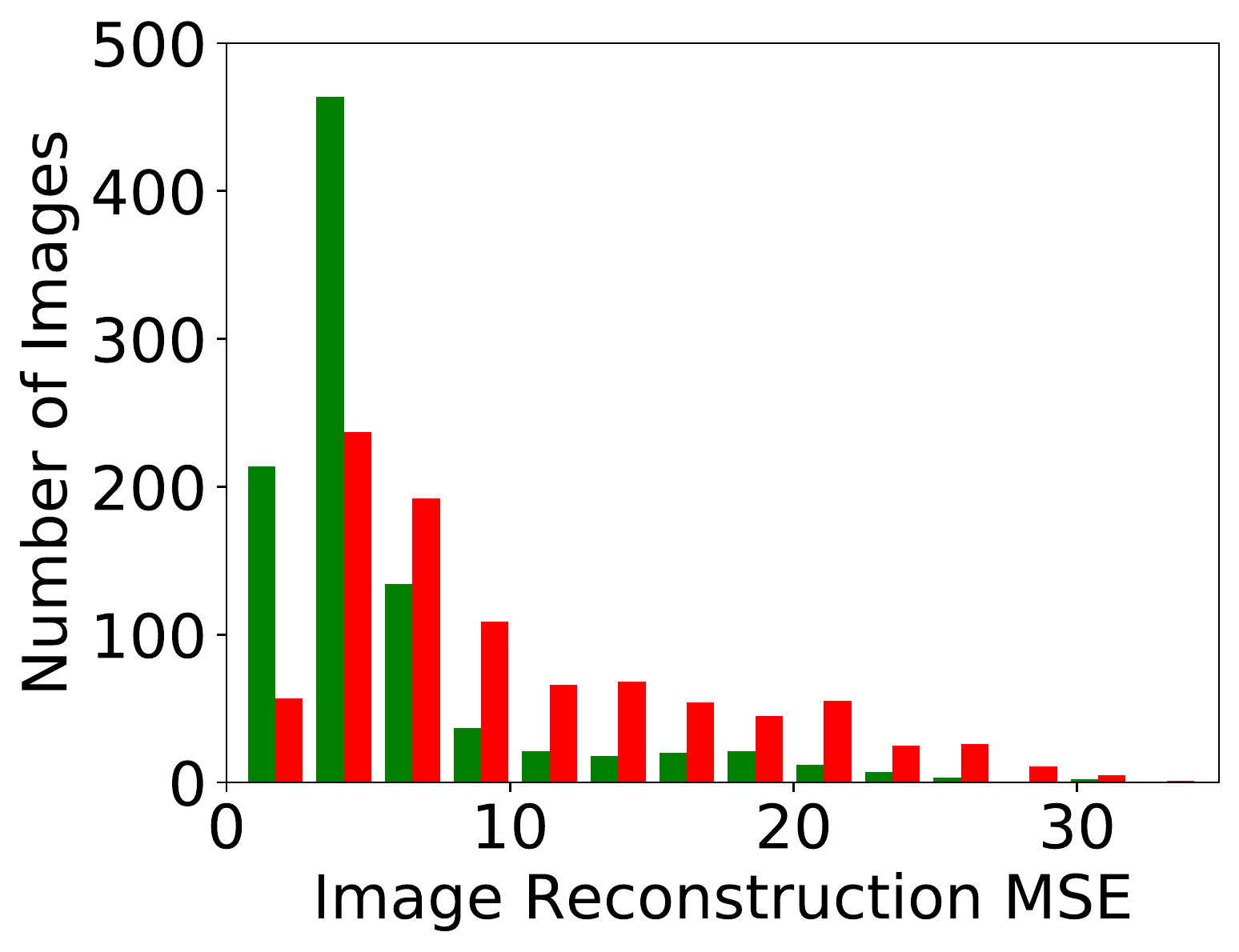}
  \caption{Mojave model, Mojave test}
  \label{fig:hist-sem4}
\end{subfigure}
\caption{Histograms of MSE within labeled ground region (\textit{green}) and vegetation region (\textit{red}) for Arroyo and Mojave test sets using Resnet18 and 256 bottleneck size.}
\label{fig:histograms-semantic}
\end{figure}

\begin{figure*}[!h]
\centering
\begin{subfigure}{0.32\textwidth}
  \centering
  \includegraphics[width=\linewidth]{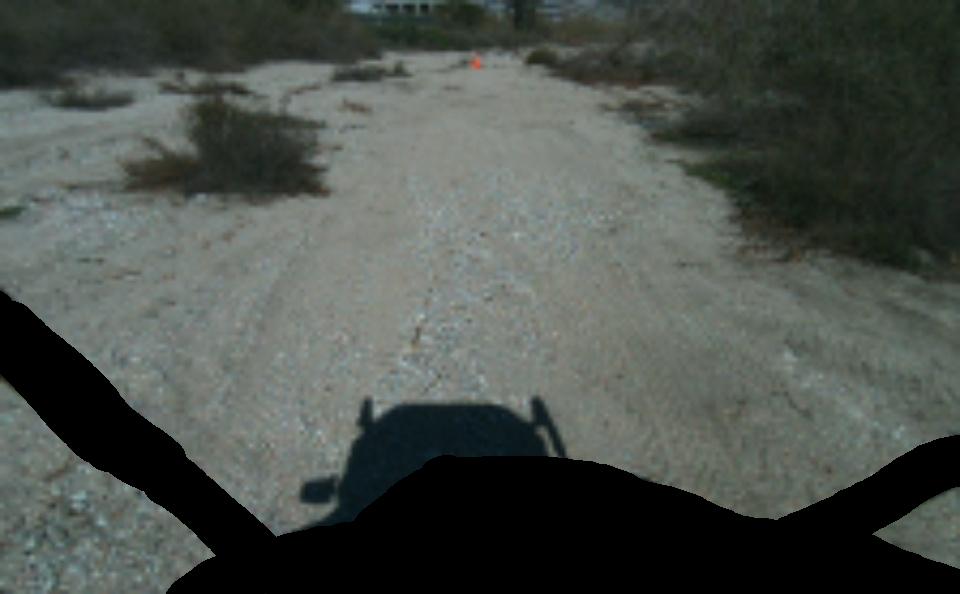}
  \label{fig:autosamp1}
\end{subfigure}
\begin{subfigure}{0.32\textwidth}
  \centering
  \includegraphics[width=\linewidth]{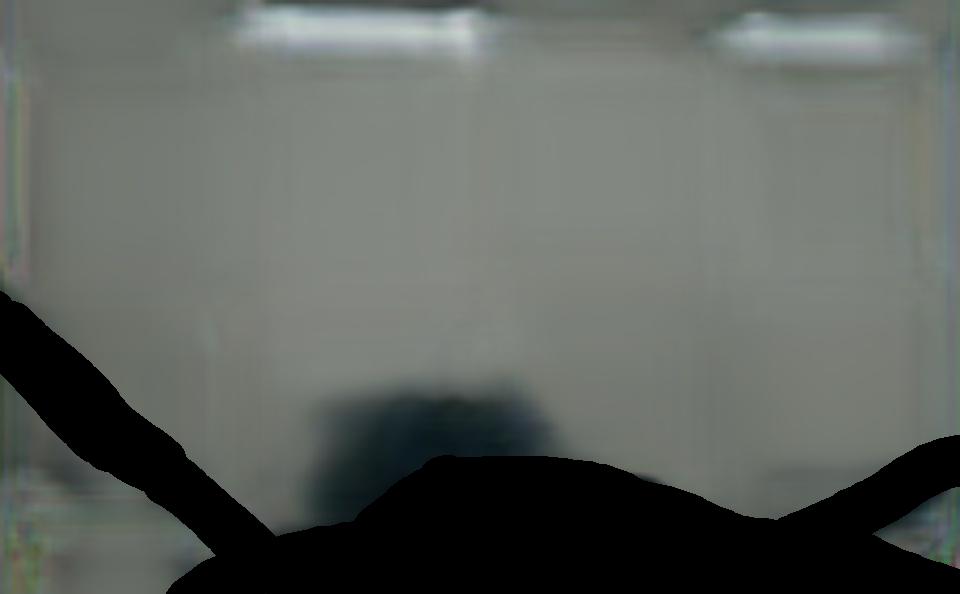}
  \label{fig:autosamp2}
\end{subfigure}
\begin{subfigure}{0.32\textwidth}
  \centering
  \includegraphics[width=\linewidth]{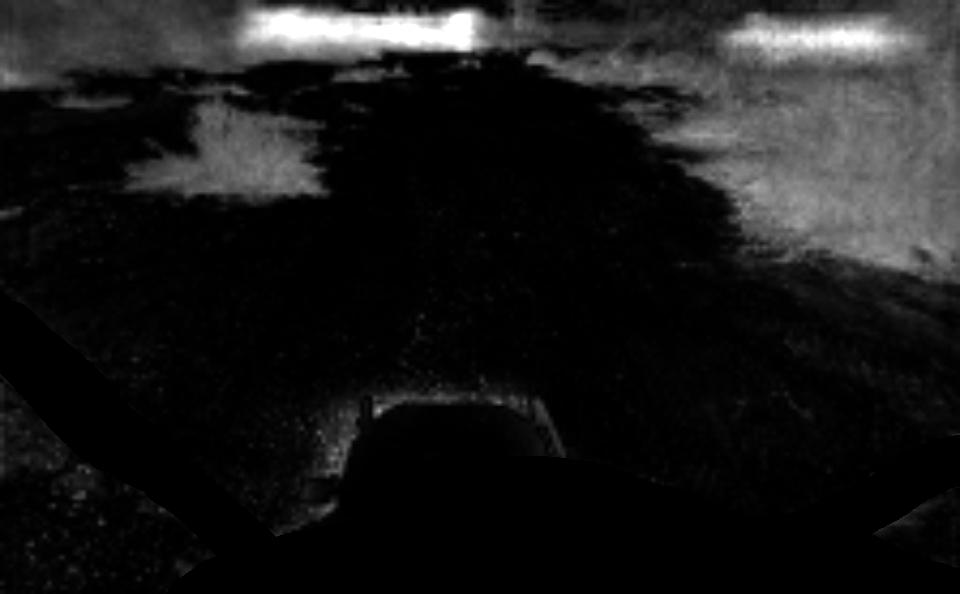}
  \label{fig:autosamp3}
\end{subfigure}

\begin{subfigure}{0.32\textwidth}
  \centering
  \includegraphics[width=\linewidth]{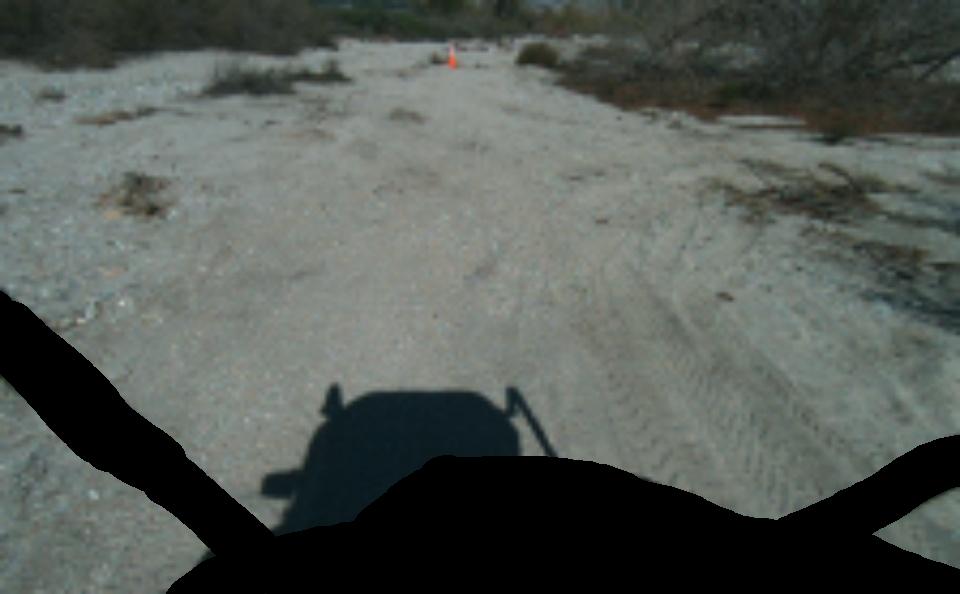}
  \label{fig:autosamp7}
\end{subfigure}
\begin{subfigure}{0.32\textwidth}
  \centering
  \includegraphics[width=\linewidth]{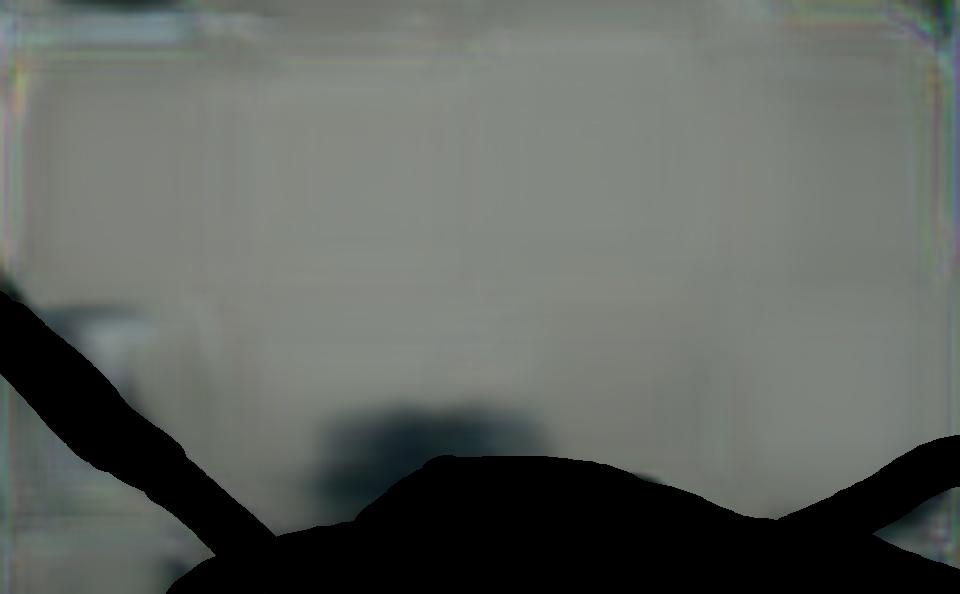}
  \label{fig:autosamp8}
\end{subfigure}
\begin{subfigure}{0.32\textwidth}
  \centering
  \includegraphics[width=\linewidth]{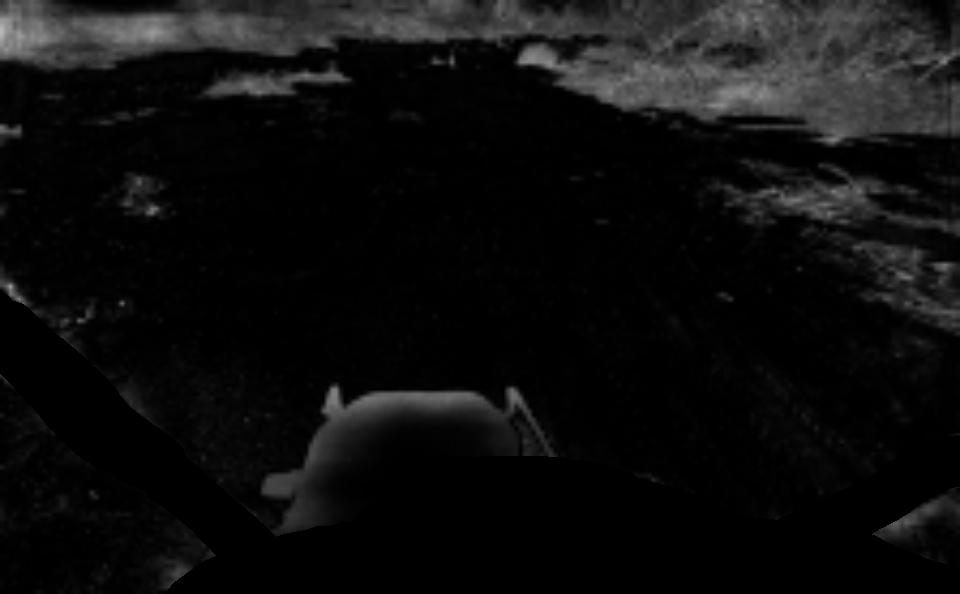}
  \label{fig:autosamp9}
\end{subfigure}

\begin{subfigure}{0.32\textwidth}
  \centering
  \includegraphics[width=\linewidth]{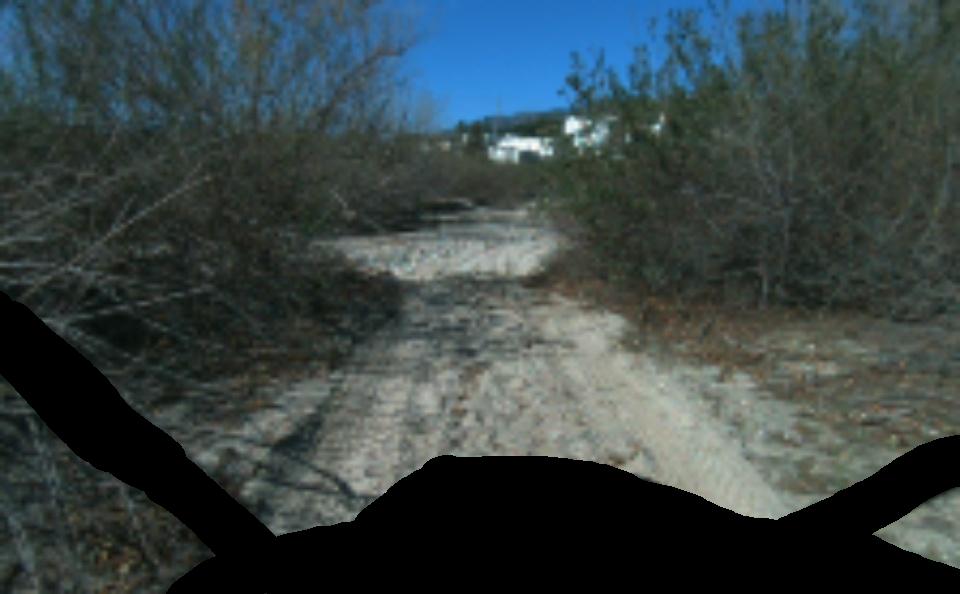}
  \label{fig:autosamp4}
\end{subfigure}
\begin{subfigure}{0.32\textwidth}
  \centering
  \includegraphics[width=\linewidth]{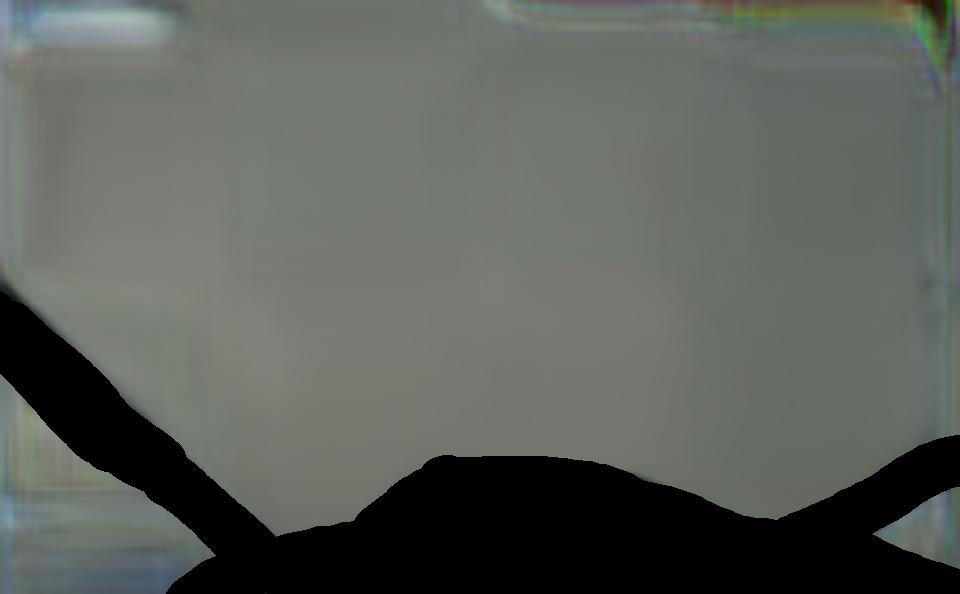}
  \label{fig:autosamp5}
\end{subfigure}
\begin{subfigure}{0.32\textwidth}
  \centering
  \includegraphics[width=\linewidth]{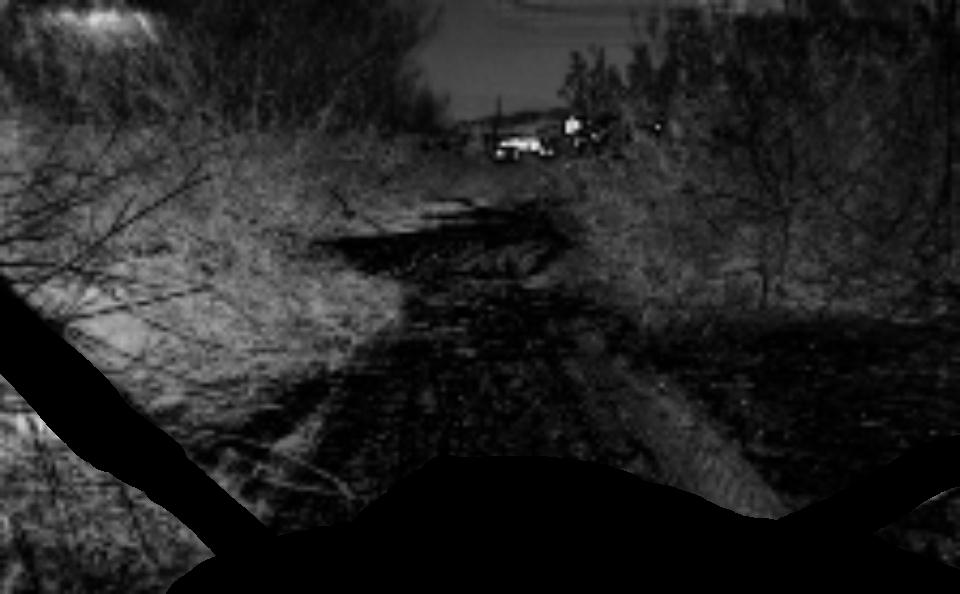}
  \label{fig:autosamp6}
\end{subfigure}

\begin{subfigure}{0.32\textwidth}
  \centering
  \includegraphics[width=\linewidth]{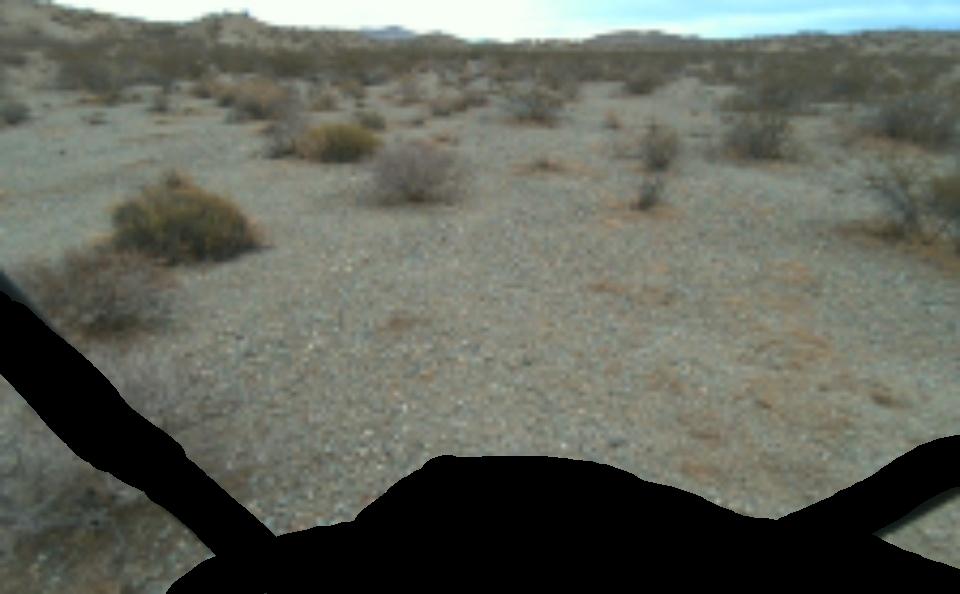}
  \label{fig:autosamp10}
\end{subfigure}
\begin{subfigure}{0.32\textwidth}
  \centering
  \includegraphics[width=\linewidth]{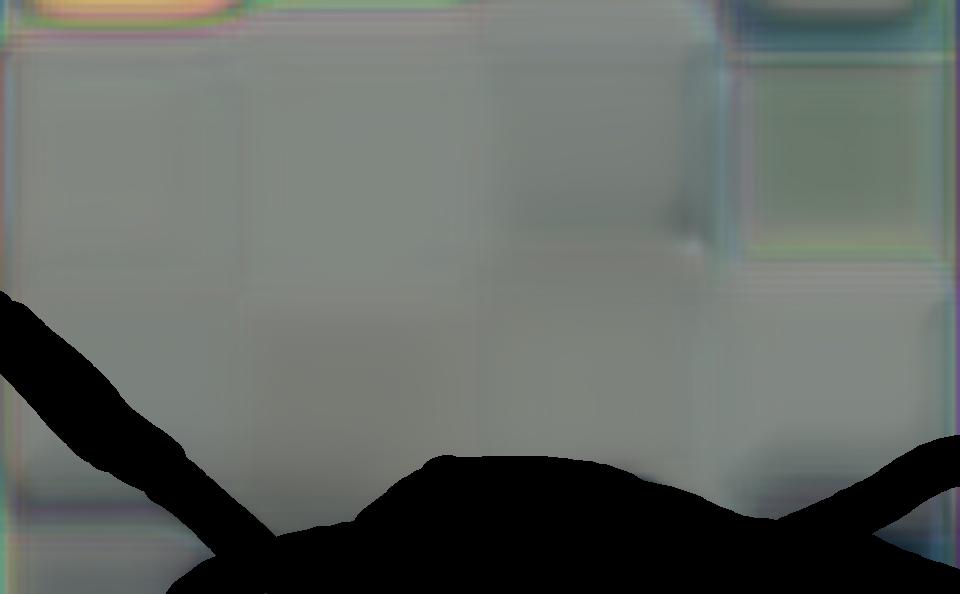}
  \label{fig:autosamp11}
\end{subfigure}
\begin{subfigure}{0.32\textwidth}
  \centering
  \includegraphics[width=\linewidth]{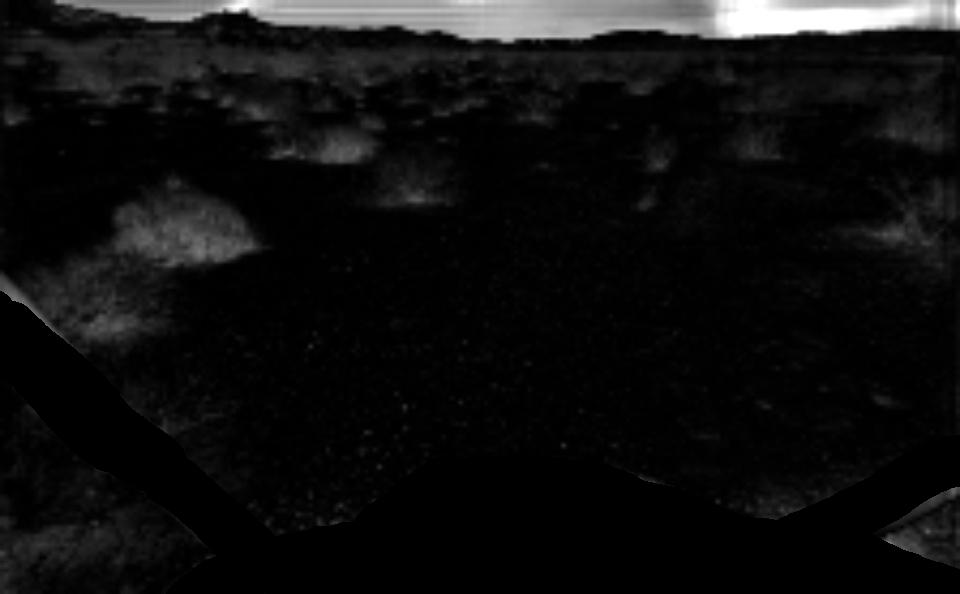}
  \label{fig:autosamp12}
\end{subfigure}
\begin{subfigure}{0.32\textwidth}
  \centering
  \includegraphics[width=\linewidth]{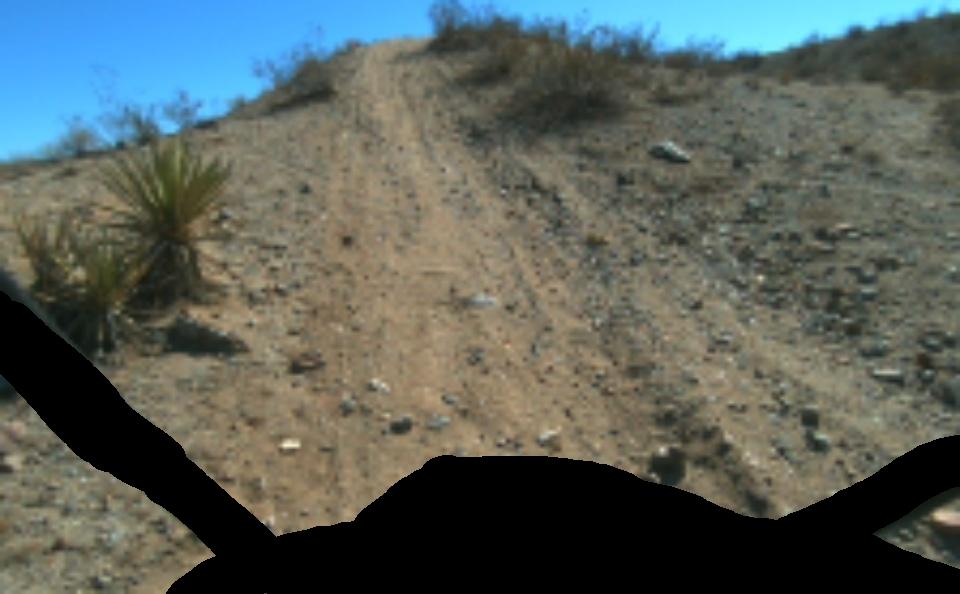}
  \caption{Original image}
  \label{fig:autosamp13}
\end{subfigure}
\begin{subfigure}{0.32\textwidth}
  \centering
  \includegraphics[width=\linewidth]{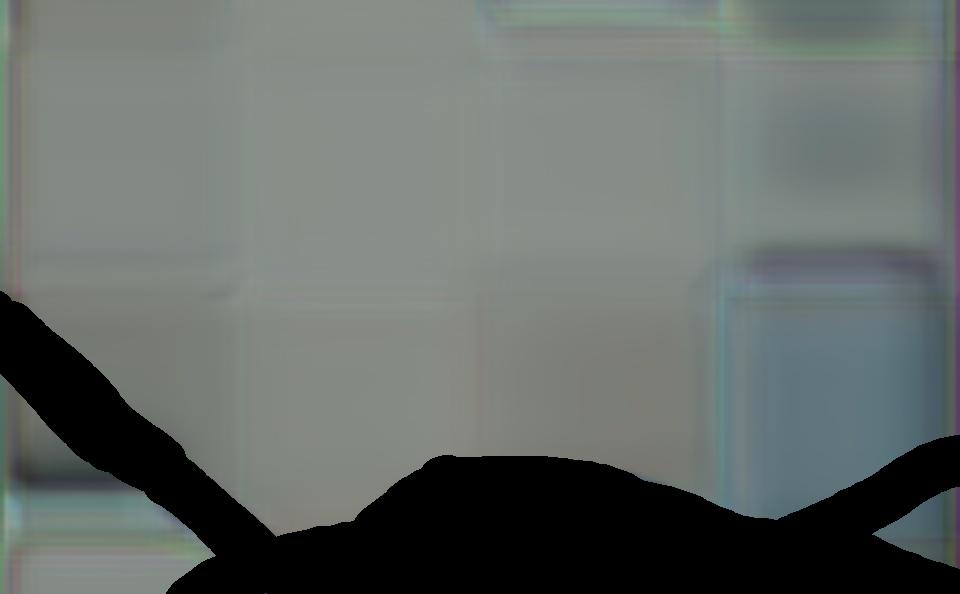}
  \caption{Reconstructed image}
  \label{fig:autosamp14}
\end{subfigure}
\begin{subfigure}{0.32\textwidth}
  \centering
  \includegraphics[width=\linewidth]{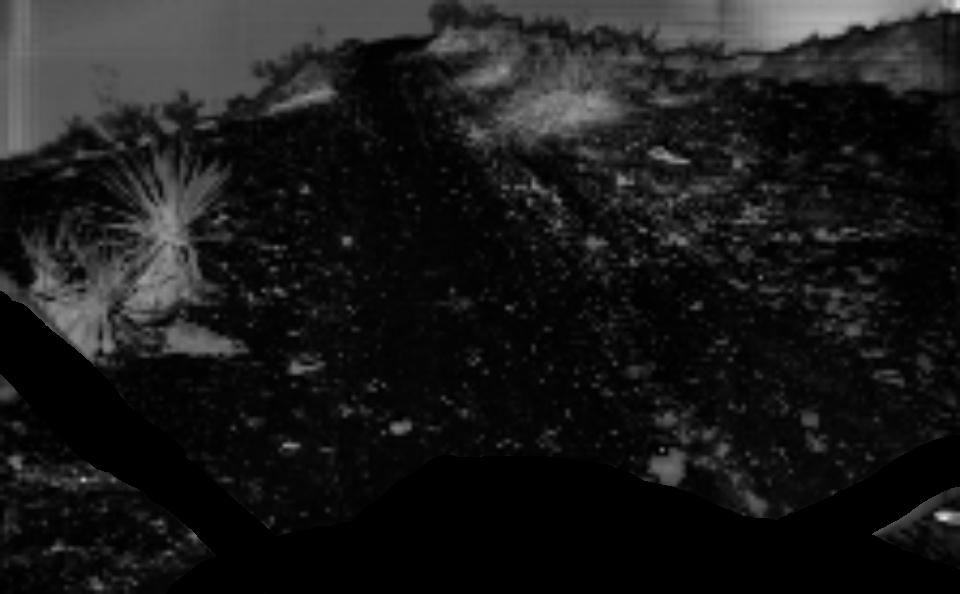}
  \caption{Reconstruction error}
  \label{fig:autosamp15}
\end{subfigure}
\caption{Sample images, reconstructions, and scaled error images for both Arroyo (\textit{top two rows}) and Mojave (\textit{bottom three rows}) data. The \textit{top two rows} were trained with Arroyo training data and \textit{bottom three rows} with Mojave training data. The \textit{first row} shows an example where the vehicle shadow is successfully reconstructed, whereas in \textit{row two} the vehicle shadow is predicted as high-risk terrain. This is one of the main limitations of the model. \textit{Row three} shows that some structures in the sand which are represented darker because of shadows are predicted with a higher risk than similar terrain. \textit{Row three} and \textit{row four} show differences of the predicted risk of traversable and non-traversable vegetation. The vegetation in \textit{row three} is large and non-traversable and is predicted with a higher risk than the small traversable vegetation in \textit{row four}. This is an interesting side effect due to the different color of the vegetation. \textit{Row five} shows that some small rocks can be predicted as high-risk.}
\label{fig:autoenc_samples}
\end{figure*}

Further analysis based on a histogram of MSE per image of the Resnet18 model with 256 bottleneck is show in Figure \ref{fig:histograms-semantic}. From this we observe that the ground often contains lower MSE compared to the vegetation. There is some overlap within these values especially within the Mojave test set. This is due to small traversable vegetation having low error, and some shadowed regions on the ground leading to a higher error. Observing the sample images, reconstructions, and scaled error images in Figure \ref{fig:autoenc_samples}, we see a good qualitative performance of the model. These samples are from a model with a Resnet18 backbone with a bottleneck size of 256. Next to shadows some small rocks and some of the texture of the sand have a large reconstruction error. The model shows a strong response for large vegetation and a smaller response for small vegetation, which is an interesting side effect of the color of the different types of vegetation.\\

\section{Conclusion and Future Work}
We present a novel pipeline with automated wheel projection and occlusion filtering which generates images to predict high and low-risk traversable terrain. We trained a variational autoencoder with a Resnet backbone and evaluated the generalizability of our model with data from a test site not seen during training. In our approach, we used the reconstruction loss as a measure of traversability. Based on our results of being able to identify 81\% of the vegetation in the Arroyo and 77\% of the vegetation in Mojave as high-risk while maintaining around 85\% of the ground as low-risk, our model was able to successfully learn to separate high and low traversability risk. This was a promising result for future usage onboard for autonomous off road driving. For a vehicle such as the Poloris RZR, which has a high risk tolerance, the lack of manual labeling offers the ability to quickly scale a perception system to many different complex environments.

In future work, we want to use our tool to not only predict traversable and non-traversable terrain but other sparse terrain properties from proprioceptive data such as speed, vibration, wheel slip, etc. Further we want to investigate techniques for better handling shadows. To further increase the performance of the model for out-of-distribution samples, negative samples can also be added. Negative samples do not necessarily need to come from driving on non-traversable areas but from driver feedback such as sudden steering or breaking in front of obstacles. 


\section{Acknowledgements}
We would like to thank Hiro Ono and the MAARS project for sharing semantic labels. The research was carried out at the Jet Propulsion Laboratory, California Institute of Technology, under a contract with the National Aeronautics and Space Administration (80NM0018D0004), partial funded by the Jet Propulsion Laboratory and partially funded by the Swiss Federal Institute of Technology, ETH Zürich. 

\bibliographystyle{bibliography/IEEEtran.bst}
\bibliography{bibliography/references}

\begin{thebibliography}{10}
\providecommand{\url}[1]{#1}
\csname url@samestyle\endcsname
\providecommand{\newblock}{\relax}
\providecommand{\bibinfo}[2]{#2}
\providecommand{\BIBentrySTDinterwordspacing}{\spaceskip=0pt\relax}
\providecommand{\BIBentryALTinterwordstretchfactor}{4}
\providecommand{\BIBentryALTinterwordspacing}{\spaceskip=\fontdimen2\font plus
\BIBentryALTinterwordstretchfactor\fontdimen3\font minus
  \fontdimen4\font\relax}
\providecommand{\BIBforeignlanguage}[2]{{%
\expandafter\ifx\csname l@#1\endcsname\relax
\typeout{** WARNING: IEEEtran.bst: No hyphenation pattern has been}%
\typeout{** loaded for the language `#1'. Using the pattern for}%
\typeout{** the default language instead.}%
\else
\language=\csname l@#1\endcsname
\fi
#2}}
\providecommand{\BIBdecl}{\relax}
\BIBdecl

\bibitem{Overbye2021GVOMAG}
T.~Overbye and S.~Saripalli, ``G-vom: A gpu accelerated voxel off-road mapping
  system,'' \emph{ArXiv}, vol. abs/2109.13176, 2021.

\bibitem{Fankhauser2018ProbabilisticTM}
P.~Fankhauser, M.~Bloesch, and M.~Hutter, ``Probabilistic terrain mapping for
  mobile robots with uncertain localization,'' \emph{IEEE Robotics and
  Automation Letters}, vol.~3, pp. 3019--3026, 2018.

\bibitem{Sofman2006ImprovingRN}
B.~Sofman, E.~Lin, J.~A. Bagnell, J.~Cole, N.~Vandapel, and A.~Stentz,
  ``Improving robot navigation through self-supervised online learning,''
  \emph{J. Field Robotics}, vol.~23, pp. 1059--1075, 2006.

\bibitem{lalonde}
J.-F. Lalonde, N.~Vandapel, D.~Huber, and M.~Hebert, ``Natural terrain
  classification using three-dimensional ladar data for ground robot
  mobility,'' \emph{J. Field Robotics}, vol.~23, pp. 839--861, 10 2006.

\bibitem{jiang2020rellis3d}
P.~Jiang, P.~Osteen, M.~Wigness, and S.~Saripalli, ``Rellis-3d dataset: Data,
  benchmarks and analysis,'' 2020.

\bibitem{yamaha}
D.~Maturana, P.-W. Chou, M.~Uenoyama, and S.~Scherer, ``Real-time semantic
  mapping for autonomous off-road navigation,'' in \emph{Field and Service
  Robotics}.\hskip 1em plus 0.5em minus 0.4em\relax Springer, 2018, pp.
  335--350.

\bibitem{freiburgforest}
A.~Valada, G.~Oliveira, T.~Brox, and W.~Burgard, ``Deep multispectral semantic
  scene understanding of forested environments using multimodal fusion,'' in
  \emph{International Symposium on Experimental Robotics (ISER)}, 2016.

\bibitem{Wellhausen2019WhereSI}
L.~Wellhausen, A.~Dosovitskiy, R.~Ranftl, K.~Walas, C.~Cadena, and M.~Hutter,
  ``Where should i walk? predicting terrain properties from images via
  self-supervised learning,'' \emph{IEEE Robotics and Automation Letters},
  vol.~4, pp. 1509--1516, 2019.

\bibitem{Zurn2021SelfSupervisedVT}
J.~Zurn, W.~Burgard, and A.~Valada, ``Self-supervised visual terrain
  classification from unsupervised acoustic feature learning,'' \emph{IEEE
  Transactions on Robotics}, vol.~37, pp. 466--481, 2021.

\bibitem{Nava2019LearningLP}
M.~Nava, J.~Guzzi, R.~O. Chavez-Garcia, L.~M. Gambardella, and A.~Giusti,
  ``Learning long-range perception using self-supervision from short-range
  sensors and odometry,'' \emph{IEEE Robotics and Automation Letters}, vol.~4,
  pp. 1279--1286, 2019.

\bibitem{Papadakis2013TerrainTA}
P.~Papadakis, ``Terrain traversability analysis methods for unmanned ground
  vehicles: A survey,'' \emph{Eng. Appl. Artif. Intell.}, vol.~26, pp.
  1373--1385, 2013.

\bibitem{MAARS}
M.~Ono, B.~Rothrock, K.~Otsu, S.~Higa, Y.~Iwashita, A.~Didier, T.~Islam,
  C.~Laporte, V.~Sun, K.~Stack, J.~Sawoniewicz, S.~Daftry, V.~Timmaraju,
  S.~Sahnoune, C.~A. Mattmann, O.~Lamarre, S.~Ghosh, D.~Qiu, S.~Nomura, H.~Roy,
  H.~Sarabu, G.~Hedrick, L.~Folsom, S.~Suehr, and H.~Park, ``Maars: Machine
  learning-based analytics for automated rover systems,'' in \emph{2020 IEEE
  Aerospace Conference}, 2020, pp. 1--17.

\bibitem{Fankhauser2016AUG}
P.~Fankhauser and M.~Hutter, ``A universal grid map library: Implementation and
  use case for rough terrain navigation,'' 2016.

\bibitem{deeplab}
\BIBentryALTinterwordspacing
L.~Chen, G.~Papandreou, I.~Kokkinos, K.~Murphy, and A.~L. Yuille, ``Deeplab:
  Semantic image segmentation with deep convolutional nets, atrous convolution,
  and fully connected crfs,'' \emph{CoRR}, vol. abs/1606.00915, 2016. [Online].
  Available: \url{http://arxiv.org/abs/1606.00915}
\BIBentrySTDinterwordspacing

\bibitem{RUGD2019IROS}
M.~Wigness, S.~Eum, J.~G. Rogers, D.~Han, and H.~Kwon, ``A rugd dataset for
  autonomous navigation and visual perception in unstructured outdoor
  environments,'' in \emph{International Conference on Intelligent Robots and
  Systems (IROS)}, 2019.

\bibitem{ai4mars}
R.~M. Swan, D.~Atha, H.~A. Leopold, M.~Gildner, S.~Oij, C.~Chiu, and M.~Ono,
  ``Ai4mars: A dataset for terrain-aware autonomous driving on mars,'' in
  \emph{2021 IEEE/CVF Conference on Computer Vision and Pattern Recognition
  Workshops (CVPRW)}, 2021, pp. 1982--1991.

\bibitem{freiburgself}
A.~Valada, R.~Mohan, and W.~Burgard, ``Self-supervised model adaptation for
  multimodal semantic segmentation,'' \emph{International Journal of Computer
  Vision (IJCV)}, jul 2019, special Issue: Deep Learning for Robotic Vision.

\bibitem{Otsu2016AutonomousTC}
K.~Otsu, M.~Ono, T.~J. Fuchs, I.~A. Baldwin, and T.~Kubota, ``Autonomous
  terrain classification with co- and self-training approach,'' \emph{IEEE
  Robotics and Automation Letters}, vol.~1, pp. 814--819, 2016.

\bibitem{Bai2019ThreeDimensionalVT}
C.~Bai, J.~Guo, and H.~Zheng, ``Three-dimensional vibration-based terrain
  classification for mobile robots,'' \emph{IEEE Access}, vol.~7, pp.
  63\,485--63\,492, 2019.

\bibitem{Barnes2017FindYO}
D.~Barnes, W.~P. Maddern, and I.~Posner, ``Find your own way: Weakly-supervised
  segmentation of path proposals for urban autonomy,'' \emph{2017 IEEE
  International Conference on Robotics and Automation (ICRA)}, pp. 203--210,
  2017.

\bibitem{Wellhausen2020SafeRN}
L.~Wellhausen, R.~Ranftl, and M.~Hutter, ``Safe robot navigation via
  multi-modal anomaly detection,'' \emph{IEEE Robotics and Automation Letters},
  vol.~5, pp. 1326--1333, 2020.

\bibitem{Kerner2020ComparisonON}
H.~R. Kerner, K.~L. Wagstaff, B.~D. Bue, D.~F. Wellington, S.~R. Jacob,
  P.~Horton, J.~F. Bell, C.~Kwan, and H.~B. Amor, ``Comparison of novelty
  detection methods for multispectral images in rover-based planetary
  exploration missions,'' \emph{Data Mining and Knowledge Discovery}, pp. 1 --
  34, 2020.

\bibitem{Stocco2020MisbehaviourPF}
A.~Stocco, M.~Weiss, M.~Calzana, and P.~Tonella, ``Misbehaviour prediction for
  autonomous driving systems,'' \emph{2020 IEEE/ACM 42nd International
  Conference on Software Engineering (ICSE)}, pp. 359--371, 2020.

\bibitem{Gu2020ANL}
X.~Gu, Y.~Han, and J.~Yu, ``A novel lane-changing decision model for autonomous
  vehicles based on deep autoencoder network and xgboost,'' \emph{IEEE Access},
  vol.~8, pp. 9846--9863, 2020.

\bibitem{Shan2020LIOSAMTL}
T.~Shan, B.~Englot, D.~Meyers, W.~Wang, C.~Ratti, and D.~Rus, ``Lio-sam:
  Tightly-coupled lidar inertial odometry via smoothing and mapping,''
  \emph{2020 IEEE/RSJ International Conference on Intelligent Robots and
  Systems (IROS)}, pp. 5135--5142, 2020.

\bibitem{resnet}
\BIBentryALTinterwordspacing
K.~He, X.~Zhang, S.~Ren, and J.~Sun, ``Deep residual learning for image
  recognition,'' \emph{CoRR}, vol. abs/1512.03385, 2015. [Online]. Available:
  \url{http://arxiv.org/abs/1512.03385}
\BIBentrySTDinterwordspacing

\bibitem{Fan2021STEPST}
D.~Fan, K.~Otsu, Y.~Kubo, A.~Dixit, J.~W. Burdick, and A.~akbar Agha-mohammadi,
  ``Step: Stochastic traversability evaluation and planning for safe off-road
  navigation,'' \emph{ArXiv}, vol. abs/2103.02828, 2021.

\end{thebibliography}
\addcontentsline{toc}{chapter}{Bibliography}

\end{document}